\documentclass[10pt, conference]{IEEEtran}
\IEEEoverridecommandlockouts
\usepackage{cite}
\usepackage{bm}
\usepackage{url}
\usepackage{multirow}
\usepackage{subfigure}
\usepackage{booktabs}
\usepackage[ruled,vlined,linesnumbered]{algorithm2e}
\usepackage{amsmath,amssymb,amsfonts}
\usepackage{algorithmic}
\usepackage{multirow}
\usepackage{graphicx}
\usepackage{textcomp}
\usepackage{xcolor}
\def\BibTeX{{\rm B\kern-.05em{\sc i\kern-.025em b}\kern-.08em
    T\kern-.1667em\lower.7ex\hbox{E}\kern-.125emX}}
\begin{document}

% \title{Sample Selection for Efficient and Robust \\ Federated Learning
% }
\title{FedSDG-FS: Efficient and Secure Feature Selection for Vertical Federated Learning}

\author{
Anran Li, 
Hongyi Peng, 
Lan Zhang, 
Jiahui Huang,
		Qing Guo, Han Yu, Yang Liu
}
% \thanks{
% Anran Li
% }

% \author{
% \IEEEauthorblockN{
% Anran Li\IEEEauthorrefmark{1}\textsuperscript{\textsection}, 
% Hongyi Peng\IEEEauthorrefmark{1}\IEEEauthorrefmark{8}\textsuperscript{\textsection}, 
% Lan Zhang\IEEEauthorrefmark{2}\IEEEauthorrefmark{6}, 
% Jiahui Huang\IEEEauthorrefmark{2},
% 		Qing Guo\IEEEauthorrefmark{3}, Han Yu\IEEEauthorrefmark{1}, Yang Liu\IEEEauthorrefmark{5}
%   }
% 	\IEEEauthorblockA{
%  \IEEEauthorrefmark{1}Nanyang Technological University, Singapore,
%  \IEEEauthorrefmark{8}Alibaba-NTU Singapore Joint  Research Institute \& Alibaba Group\\
% 	\IEEEauthorrefmark{2}University of Science and Technology of China, China,
%  \IEEEauthorrefmark{3}Center for Frontier AI Research (CFAR), A*STAR, Singapore\\
%  \IEEEauthorrefmark{5}Zhejiang Sci-Tech University, China,
%  \IEEEauthorrefmark{6}Institute of Artificial Intelligence, Hefei Comprehensive National Science Center, China
% 	}
% }

\maketitle
\begingroup\renewcommand\thefootnote{\textsection}
\footnotetext{This paper has been accepted by IEEE INFOCOM 2023.\\
Anran Li, Hongyi Peng, Han Yu and Yang Liu are with the school of Computer Science and Engineering of Nanyang Technological University, Singapore. Lan Zhang and Jiahui Huang are with the school of Computer Science and Technology of University of Science and Technology of China, China. Qing Guo is with the Center for Frontier AI Research, A*STAR, Singapore.}
\endgroup

\begin{abstract}
Vertical Federated Learning (VFL) enables multiple data owners, each holding a different subset of features about largely overlapping sets of data sample(s), to jointly train a useful global model. 
%Due to the existence of noisy or redundant local features, 
Feature selection (FS) is important to VFL. 
It is still an open research problem as existing FS works designed for VFL either assumes prior knowledge on the number of noisy features or prior knowledge on the post-training threshold of useful features to be selected, making them unsuitable for practical applications.
%existing FS approaches cannot be applied to VFL due to the inaccessibility of local data.
%existing FS approach designed for VFL performs feature selection out of the context of learning models, and relies on a trusted third party (TTP) and prior knowledge of noisy features, making it unsuitable for practical VFL systems.
To bridge this gap, we propose the Federated Stochastic Dual-Gate based Feature Selection (FedSDG-FS) approach.
It consists of a Gaussian stochastic dual-gate to efficiently approximate the probability of a feature being selected, with privacy protection through Partially Homomorphic Encryption without a trusted third-party. To reduce overhead, we propose a feature importance initialization method based on Gini impurity, which can accomplish its goals with only two parameter transmissions between the server and the clients.
%Theoretically, we analyze the convergence rate of VFL-FS. 
% It consists of a Gaussian stochastic dual-gate for clients' inputs to efficiently approximate the probability of a feature being selected. A feature selection approach, which leverages Gini impurity for VFL then initializes the importance of individual features. Important features and significant local embeddings can then be selected by the proposed stochastic dual gates. The feature importance initialization step can be accomplished by VFL-FS through a single parameter transmission between the server and the clients with a single encryption/decryption operation on the server based on Gini impurity and Partially Homomorphic Encryption (PHE), which significantly improves efficiency and privacy preservation.
%Empirically, we evaluate VFL-FS through extensive experiments on both synthetic and real datasets (including tabular data, images, texts, and audios). 
Extensive experiments on both synthetic and real-world datasets show that FedSDG-FS significantly outperforms existing approaches in terms of achieving accurate selection of high-quality features as well as building global models with improved performance. 

% , \emph{e.g.}, with the average 13.32 using 20\% fewer features. 

% It is a hierarchical feature selection approach which first distinguishes qualified features from noisy features, and then selects the most important features amongst the qualified features. The noisy feature identification step can be accomplished through a single parameter transmission between the server and the clients, and only one encryption/decryption operation on the server based on Gini impurity and Partially Homomorphic Encryption (PHE). The secure important feature selection method of VFL-FS leverages the Gaussian-based stochastic gates for VFL, and is designed based on PHE and the randomized noise mechanism  to simultaneously determine important features and construct the global model to enhance performance. 
% VFL-FS has been evaluated through extensive experiments on both synthetic and real datasets (including tabular data, images, texts, and audios). The results show that VFL-FS significantly outperforms existing approaches in terms of achieving accurate and secure selection of high-quality features as well as building high-performance global models.

\end{abstract}

\begin{IEEEkeywords}
Feature selection, vertical federated learning
\end{IEEEkeywords}

\section{Introduction}
\label{sec:intro}
% The success of artificial intelligence (AI) rests on the availability of large amounts of data. 
% However, when data has been isolated across different parties or organizations, it limits AI's capability in real-world applications.
Federated learning (FL) \cite{mcmahan2017communication, hu2019fdml, yang2019federated, wang2022efficient,li2021privacy1} is an emerging machine learning pardigm, which enables multiple data owners to jointly train a model by iteratively exchanging model parameters through an FL server, while preserving local data privacy. 
% Consequently, Google proposed a federated learning (FL) system \cite{mcmahan2017communication,hard2018federated,hu2019fdml}, in which a global machine learning (ML) model is constructed by distributed participants without exposing their local data to any third party.
Based on the distribution of local data, there are two main categories of FL scenarios: 1) horizontal federated learning (HFL) and 2) vertical federated learning (VFL).
Under HFL \cite{li2021efficient, zhuang2021joint, li2021sample}, data owners' local datasets have little overlap in the sample space but large overlaps in the feature space. 
Under VFL \cite{liu2019communication, chen2020vafl, tan2022residue}, data owners' local datasets have large overlaps in the sample space but little overlap in the feature space. VFL scenarios often arise in real applications \cite{PowerFL,FATE}, \emph{e.g.}, an e-commerce company, a bank and a ride-sharing company can collaborate to build a model to identify potential financial fraudsters based on the multiple perspectives on people's behaviour through VFL. 
% when companies from different business sectors join forces to train a machine learning model 
% Among these FL methods, VFL is the most popular one in e-commerce, financial, and healthcare applications \cite{PowerFL,FATE}, \emph{e.g.}, an e-commerce company may want to
% estimate the click-through rates (CTR) of customers using their historical transactions from multiple financial institutions. Specifically, VFL describes the scenarios where multiple parties handle data with the same sample IDs, but each party has its own feature set.
The quality of data owners' local features determines the effectiveness of their local models, thereby affecting the performance of the global VFL model. In practice, data owners can possess noisy features that are irrelevant to the learning task, or a large number of redundant features, which seriously impairs global model performance.
As an example, one of our experiments in Section \ref{sec:motivating} shows that a two-class classifier trained by VFL with the real dataset suffered an accuracy loss from 82.6\% to 54.2\% due to the existence of noisy features. 

%\footnote{As illustrated by the experiment results in Section \ref{sec:motivation}.}

% Besides, since each client has no access to other clients' features, his/her locally used ones for modeling would overlap with those of others, therefore results in excessive parameter transmissions. See the motivation examples in Section \ref{sec:motivation} for a more concrete explanation.

% The feature sets used locally for modeling determines their local models, thereby affecting the performance of the global model. 
% As an example, for large-scale recommendation systems involving large-cardinality categorical features, the input component of the VFL model assigns an embedding vector to each item of discrete features.
% This results in a huge number of embedding parameters in the input component, which dominate both the size and the inference time cost of the model.

% There are large number of local features, but usually the task at hand depends only on a small subset of these features \cite{yamada2020feature}. 
% In practice, however, parties often use all local features to train the VFL model regardless of the complexity of the task.
% As an example, for large-scale recommendation system involving large-cardinality categorical features, the input component of the VFL model assigns an embedding vector to each item of discrete features, which results in a huge number of embedding parameters in the input component and the global model.

To improve the performance of VFL systems, in this work, we focus on filtering noisy features and selecting important features. 
A number of feature selection methods have been proposed for centralized machine learning settings \cite{li2021privacy,yamada2020feature,chen2017kernel}, while few work focused on VFL \cite{pansecure}.
Feature selection methods for centralized machine learning can be divided into three categories: 1) \textbf{filter methods} calculate per-feature relevance scores based on statistical measures (\emph{e.g.}, Gini impurity) to filter features prior to learning a model \cite{chen2017kernel,song2012feature, estevez2009normalized}; 2) \textbf{wrapper methods} search for the optimal feature subset in large search spaces \cite{roy2015feature,kabir2010new}; and 3) \textbf{embedded methods} attempt to select subset of important features while simultaneously learn the model \cite{li2016deep,yamada2020feature,hans2009bayesian}. 

%In VFL, there are only two works on feature selection (FS) \cite{pansecure, feng2022vertical}. In \cite{pansecure}, FS is performed with the filter method based on secure multi-party computation. However, since it performs VFL FS out of the context of the learning task, it can lead to inaccurate feature selection. Besides, it assumes that the number of noisy features is known in advance, and that there is a trusted third party (TTP) for performing FS. These assumptions are unrealistic in practice. Further, it incur large communication overhead since a massive amount of parameters are transmitted between participants and the TTP. In \cite{feng2022vertical} the embedded method combined the auto-encoder with $l_2$ constraints on feature weights is used for FS. However, it suffers from shrinkage of the model parameters, and requires post-training threshold setting to determine the selected features \cite{louizos2017learning}.   

% while the other \cite{feng2022vertical} utilizes the embedded method with auto-encoder for data projection. Those two methods have the following limitations. 

Existing FS works designed for VFL either assumes prior knowledge on the number of noisy features \cite{pansecure} or prior knowledge on the post-training threshold of useful features to be selected \cite{louizos2017learning}. These assumptions make them unsuitable for practical VFL applications. The problem of feature selection in VFL settings remains open. To enable feature selection to be performed in VFL settings, the following key research questions need to be addressed. 1) \emph{How to accurately identify noisy features, and select a small number of important features to train an optimal global VFL model in a privacy-preserving manner?}
Existing FS methods require direct access to training samples, the training process and the labels simultaneously, which is not permitted in VFL. Besides, during VFL training, intermediate parameters are transmitted in ciphertexts \cite{zhang2020additively, cheng2021secureboost}, which further increases the difficulty of feature selection. 2) \emph{How to conduct feature selection efficiently and adaptively in VFL settings? }
Existing FS methods require a large number of training iterations to select features, especially for high-dimensional data \cite{li2016deep,yamada2020feature}. Directly applying them in VFL will incur significant computation and communication overhead since each training round involves multiple encryption/decryption operations and intermediate parameter transfers. 

% \subsubsection{\textcolor{red}{Design Overview}}
% \textcolor{red}{Illustrate our design ideas and principles:
% }

To address the aforementioned questions and the limitations of existing works \cite{pansecure, louizos2017learning}, we propose the Federated Stochastic Dual-Gate based Feature Selection (FedSDG-FS) approach. It is an embedded feature selection approach consisting of a feature importance initialization module and a secure important feature selection module.
Its advantages are summarized as follows:
\begin{itemize}
    \item \textbf{Context-Awareness}: FedSDG-FS can jointly perform feature selection and model training following the proposed stochastic dual-gate and Gini impurity-based feature importance initialization, thereby ensuring the selected features be to relevant to the context of the model. 
    %  We propose an efficient and secure feature selection framework, which first filters out noisy features before VFL training, and then selects important features during VFL training.  
    \item \textbf{Efficiency}: The FedSDG-FS Gini impurity based feature importance initialization enables the global model to quickly filter out noisy features and select important ones, thus speeding up model training. 
    The stochastic dual-gates are designed to reduce the sizes of the embedding vectors, thereby saving communication costs. 
    
    \item \textbf{Security}: FedSDG-FS achieves secure feature selection and model training by leveraging partially homomorphic encryption (PHE) and the randomized mechanism. During the feature selection and model training process, neither data nor labels are exposed to any party other than their original owners. 
    
%    \item A Gaussian stochastic dual-gates for VFL is designed for clients' inputs to efficiently approximate the probability of a feature being selected. We propose a secure feature selection method for specific VFL tasks with the proposed stochastic dual gates as well as the PHE, \emph{e.g.}, Paillier, and randomized noise mechanism. To save overhead and facilitate selection, we propose an feature importance initialization method based on Gini impurity and PHE. It can be accomplished through a single parameter transmission between the server and the clients, and a single encryption/decryption operation on the server. During the entire feature selection and model training process, neither data nor labels are exposed to any party other than their original owners. 
\end{itemize} 
We evaluate FedSDG-FS via extensive experiments using nine datasets including tabular data, images, texts and audios on a VFL system. The results show that it significantly outperforms existing approaches in terms of achieving accurate and secure selection of high-quality features to build high-performance VFL models. Taking \texttt{MADELON} dataset as an instance, the average test accuracy of FedSDG-FS is 27.0\% higher than that of the best performing baseline with 47\% fewer features required, and only half the communication cost.

\section{Related Works}
\label{sec:related}

Feature selection plays an important role in machine learning tasks. There are a number of feature selection methods proposed for centralized machine learning settings \cite{li2021privacy,yamada2020feature,chen2017kernel}, while few works deal with feature selection in VFL \cite{pansecure}.

\subsection{Feature Selection in Centralized Learning}
Feature selection methods in centralized learning settings can be divided into three categories: 1) filter methods, 2) wrapper methods, and 3) embedded methods.
Filter FS methods attempt to remove irrelevant features prior to learning a model. These methods filter features using relevance scores (\emph{e.g.}, Gini impurity) and mutual information, which are calculated based on statistical measures \cite{chen2017kernel, song2012feature, song2007supervised, li2021privacy}. 
% Usually filter methods have much less computational complexity compared to wrapper and embedded approaches.
Wrapper FS methods leverage the outcomes of a model to determine the importance of each feature. They attempt to select a subset of features which can achieve the best prediction performance. As the number of subsets can be very large in the context of deep neural networks, and a model need to be recomputed for each subset, wrapper methods are generally computationally expensive \cite{allen2013automatic, roy2015feature, kabir2010new}.
Embedded FS methods aim to select a subset of relevant features, while simultaneously learning the model \cite{li2016deep,yamada2020feature,hans2009bayesian}. 
The least absolute shrinkage and selection operator  \cite{hans2009bayesian} is a well-known embedded FS method, whose objective is to minimize the loss while enforcing an $l_1$ constraint on the weights of the features.
Another recently proposed method \cite{yamada2020feature} uses a continuously relaxed Bernoulli variable to conduct FS based on stochastic gates. 
However, this method requires a large number of parameters to be trained in the first layer, resulting in overfitting to the training data, especially for deep neural networks with high-dimensional data or when there are only a limited number of training samples available. 

Since these methods are designed for centralized learning scenarios in which all training data are accessible, such approaches are not applicable to VFL which demands data privacy protection. In addition, they are also not optimized to reduce communication or computation costs when the volume of training data is large. 

\subsection{Feature Selection in VFL}
In VFL, there are only two works on feature selection (FS) \cite{pansecure, feng2022vertical}. In \cite{pansecure}, FS is performed with the filter method based on secure multi-party computation. However, since it performs VFL feature selection out of the context of the learning task, it can lead to inaccurate feature selection. Besides, it assumes that the number of noisy features is known in advance, and that there is a trusted third party for performing FS. These assumptions are unrealistic in practice. Further, it incur large communication overhead since a massive amount of parameters are transmitted between participants and the trusted third party.
In \cite{feng2022vertical} the embedded method combined the auto-encoder with $l_2$ constraints on feature weights is used for FS. However, it suffers from shrinkage of the model parameters, and requires post-training threshold setting to determine the selected features \cite{louizos2017learning}.   
The proposed FedSDG-FS approach addresses these limitations of the state of the art.

%To this end, the lack of practical solution motivates us to design a secure and efficient method to select a small number of important features and construct a high-performance global model in VFL settings.
% \subsection{Vertical Federated Learning}

% The local information of client $m$ is fully captured in the embedding vector $h_{n,m}$, $n\in[N]$. Hence, the parameters that will be exchanged between the server and clients are $\{h_{n,m}\}$ and the gradients of $l(\theta_0, h_{n,1}, \cdots, h_{n,M};y_n)$ with respect to (w.r.t.) $\{h_{n,m}\}$.

\section{Preliminaries \& Problem Definition}
\label{sec:definition}
\label{sec:motivation}
\subsection{Basic Setup of VFL}
\label{sec:intro-VFL}
There are two types of entities involved in VFL: a server $S$ and $M$ clients $\mathcal{M}:=\{1, 2, \cdots, M\}$.
A dataset $U=\{U_1, \cdots, U_M\}$ of $N$ samples, $\{X, Y\}:=\{x_n, y_n\}_{n=1}^N$, is maintained by the $M$ clients.
Let $[N]=\{1,2,\cdots,N\}$.
Each client $m$ is associated with a unique set of features $\{f_{m,1}, \cdots, f_{m,d_m}\}$, and owns sample $x_{n,m}\in \mathbb{R}^{d_m}, n\in [N]$, where $x_{n,m}$ is the $m$-th block of the $n$-th sample vector $x_n := \left[x_{n,1}^\top, x_{n,2}^\top, \cdots, x_{n,M}^\top\right]^\top$.
Suppose there are $c$ possible class labels, the $n$-th label $y_n\in [c]$ is stored by server $S$. Typically, a data owner, which holds both the feature and the class labels, can act as the ``FL server''. It is referred to as the active party. Others which hold only features are referred to as the passive parties.

Each client $m$ learns a local embedding $h_m$ parameterized by $\theta_m \in \Theta$ that maps a high-dimensional vector $x_{n,m}\in \mathbb{R}^{d_m}$ into a low-dimensional one $h_{n,m}:=h_m(\theta_m; x_{n,m})\in \mathbb{R}^{\underline{d_m}}$ with $\underline{d_m}\ll d_m$.
The server $S$ learns the prediction $\hat{y}_n$ parameterized by the top model $\theta_0:=\{w_1, \cdots, w_M, \alpha_0\}\in \Theta$, $w_m\in \mathbb{R}^{\underline{d_m}}$, $m\in[M]$, 
where $\{w_1, \cdots, w_M\}$ are parameters of the interactive layer which concatenates embedding vectors $h_{n,1}, \cdots, h_{n,M}$ in a weighted manner. $\alpha_0$ denotes the parameters of the succeeding layers of the top model connected to the interactive layer.
% subsequent layers 
% into the vector $[u_{n,1}, \cdots, u_{n,M}]$ with $u_{n,m}:=w_m\odot h_{n,m}$, $\odot$ denote the point-wise product, and $\alpha_0$ is the parameter of the top model's subsequent layers. 
% Let $h_n=[h_{n,1}, h_{n,2}, \cdots, h_{n,M}]$ concatenate embedding vectors $h_{n,1}, h_{n,2}, \cdots, h_{n,M}$ from local clients $m\in[M]$.
Ideally, the objective of VFL is to minimize, 
% construct the global model $\hat{\theta}$ that, 
% \begin{equation}
% \label{eq:vfl-goal}
% \begin{split}
%     R(\theta_0, \bm{\theta}):=\frac{1}{N}\sum_{n=1}^N l(\theta_0, h_{n,1}, \cdots, h_{n,M};y_n)\\
%     {\rm with} \,\, h_{n,m}:=h_m(\theta_m; x_{n,m}), m\in[M],
% \end{split}
% \end{equation}
\begin{equation}
\label{eq:vfl-goal}
\begin{split}
R(\bm{\theta})&:=\mathbb{E}_{X,Y} L(h(\theta_0, h_{n,1}, \cdots, h_{n,M});y_n)\\
    &{\rm with} \,\, h_{n,m}:=h_m(\theta_m; x_{n,m}), m\in[M]
\end{split}
\end{equation}
where $\bm{\theta}:=\{\theta_i\}_{i=0}^M$ denotes the global model, which consists of $M$ local models $\theta_1, \cdots, \theta_M$ and the top model $\theta_0$, and $L(\cdot;\cdot)$ is the loss function. 
This problem can be solved via iterative stochastic optimization. 
In the $t$-th iteration, the server receives embedding vectors $\{h_{n,m}^t\}_{m=1}^M$ from $M$ clients. It then calculates and sends the gradients of the loss w.r.t. $h_{n,m}^t$ to all clients.
Upon receiving the gradients, client $m$ updates the local model to obtain $\theta_m^{t+1}$. Then, client $m$ randomly selects a datum $x_{n,m}$, calculates $h_{n,m}^{t+1}$ using $\theta_m^{t+1}$, and uploads it to the server. This process is repeated until the global model converges (\emph{i.e.}, a convergence criterion is met). To ensure that neither data nor labels can be obtained or inferred by any other party, the above iterative training must be conducted in a privacy-preserving manner. 

\begin{figure}[t!]
		\centering
		\subfigure[Accuracy and model size vs. number of participating features ]{\label{fig:Arcene_motivation}
	\includegraphics[width = 0.46\linewidth, height=1.18in, trim=4 4 4 4]{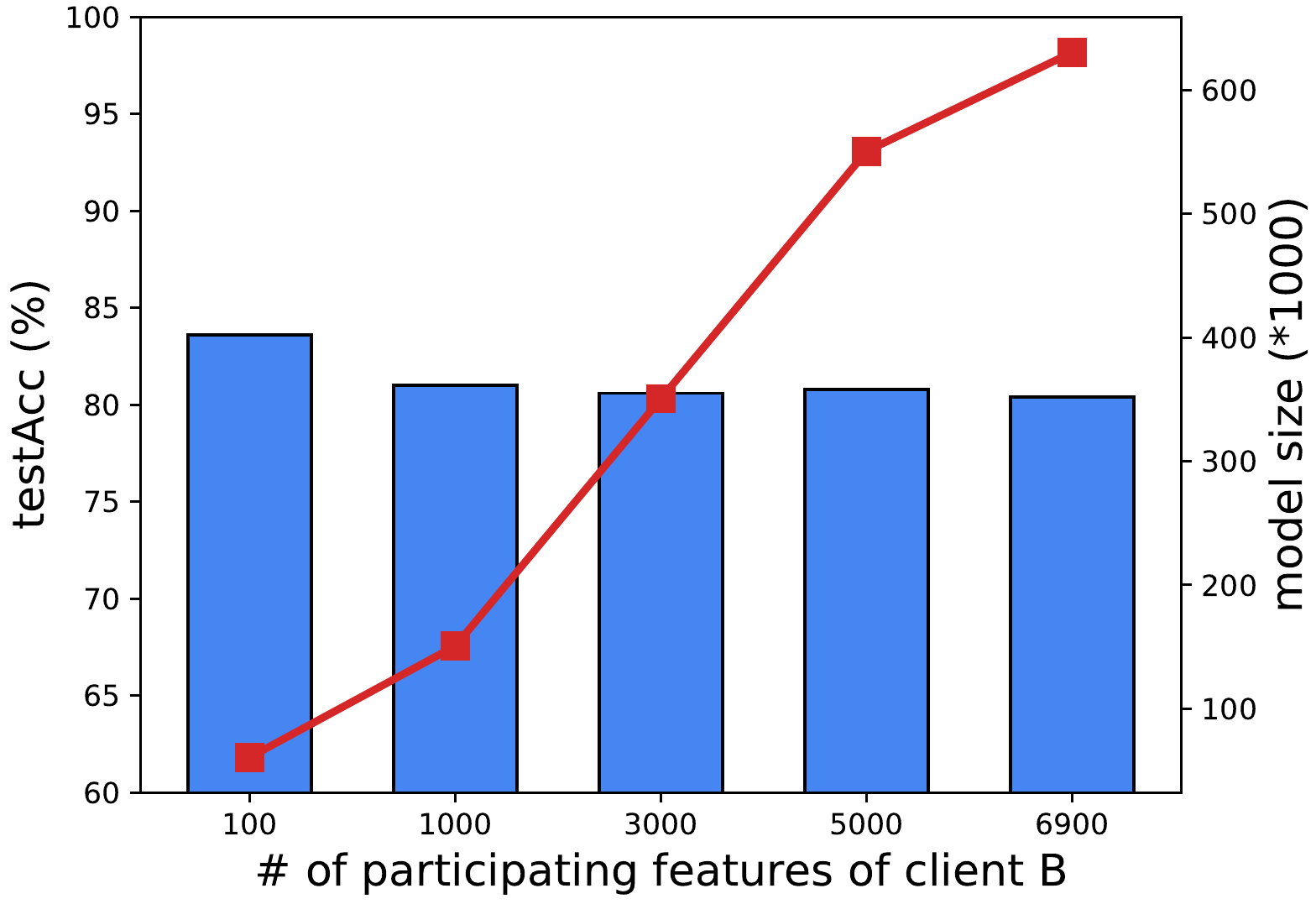}}
	\quad
	\subfigure[Accuracy vs. number of participating features]{\label{fig:madelon_motivation}
	\includegraphics[width = 0.45\linewidth, height=1.2in, trim=4 4 4 4]{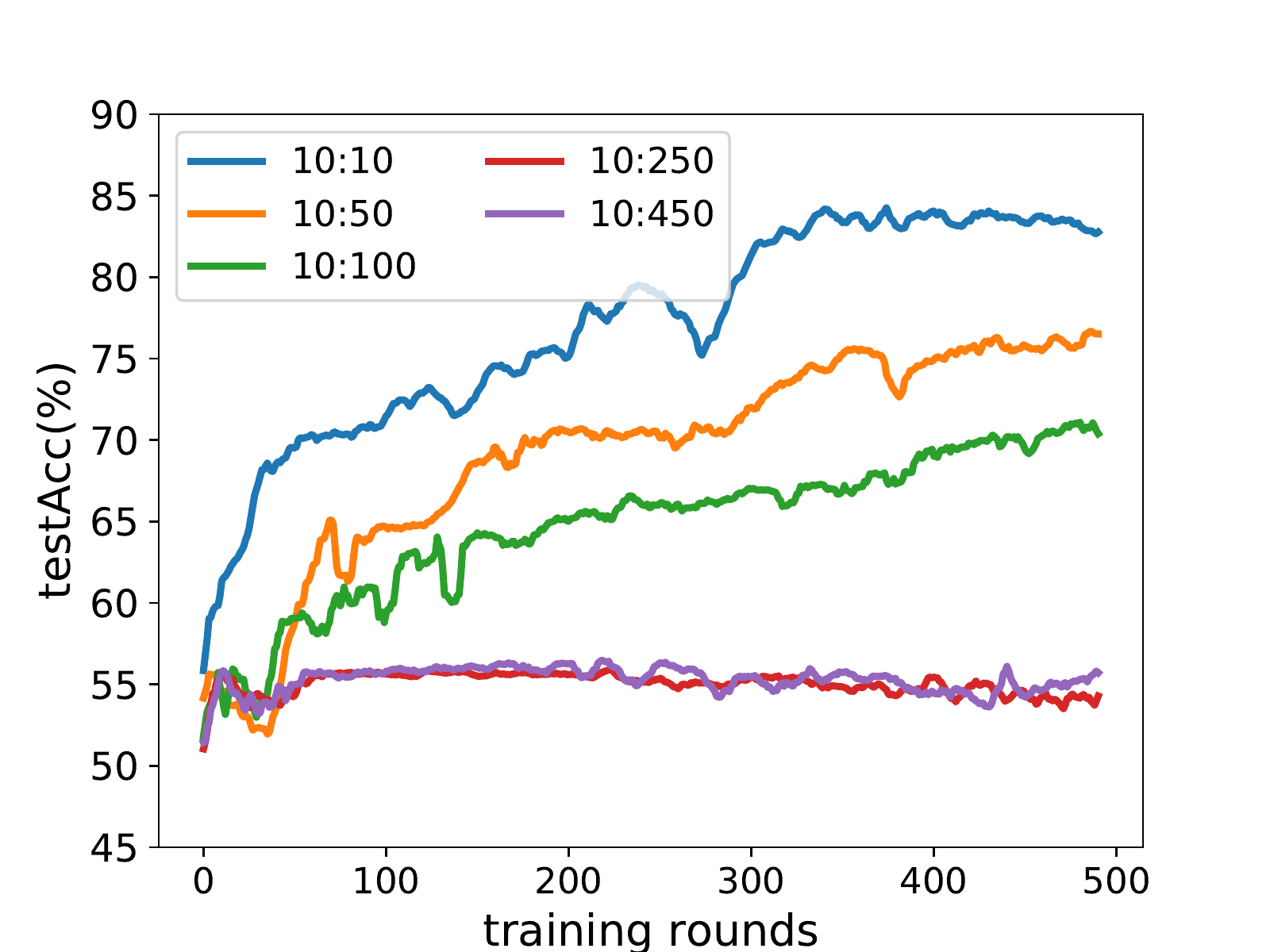}}
    	\vspace{-0.08in}
    	\caption{
	Test accuracy and model size of vertical neural networks training using (a) the dataset \texttt{ARECENE} with redundant features; 
	(b) the dataset \texttt{MADELON} with noisy features.}
	\label{fig:problem_motivation}
\vspace{-0.1in}
\end{figure}
\subsection{Motivating Examples}
\label{sec:motivating}
Here, we perform data driven analysis to demonstrate the necessity of feature selection in VFL. We illustrate this from two aspects: 1) many clients may possess a large number of redundant features, which results in a low quality and very complex global model; and 2) some clients can possess noisy or task irrelevant features which reduce global model performance. 
Specifically, we use datasets \texttt{ARCENE} \cite{ARCENE} and  \texttt{MADELON} \cite{guyon2004result} as training data to investigate the two observations.  \texttt{ARCENE} contains 2,400 instances with 7,000 informative but redundant features. \texttt{MADELON} contains 4,400 instances with 5 informative features and 480 noisy features.
We employ two clients, A and B, and a server to jointly train neural networks \cite{zhang2020additively} based on these two datasets via VFL. 

To illustrate aspect 1), we assign different numbers of features of  \texttt{ARCENE} to client B, while assigning 100 fixed features to client A to train the VFL network.
The results in Fig \ref{fig:Arcene_motivation} show that, as the number of redundant features increases, the test accuracy of the global model decreases slightly, while the model size grows rapidly. 
To illustrate aspect 2), we assign different numbers of noisy features from the \texttt{MADELON} dataset to client B, while assigning 10 fixed features to client A to train the VFL models. The results are shown in Fig. \ref{fig:madelon_motivation}, where $10:k$ indicates that, A owns $10$ features (i.e., $3$ informative features and $7$ noisy features), and B owns $k$ features (i.e., $2$ informative features and $(k-2)$ noisy features). The result shows that as the number of noisy features increases, the test accuracy of the global VFL model decreases significantly. 
These results show that an efficient and privacy-preserving feature selection method is urgently needed for VFL.

% \begin{figure}[t!]
% 	\begin{center}
% 		\includegraphics[width=\linewidth,clip]{system-overview.pdf}
% 		\caption{System overview.}
% 		\vspace{-0.1in}
% 		\label{fig:system_overview1}
% 	\end{center}
% \end{figure}

\subsection{Problem Formulation}
In a typical VFL system, under the coordination of the server $S$, all participants train a global model by transferring their local embedding vectors trained using their local datasets. 
Additionally, we consider a situation in practice in which some clients possess a large number of noisy features or redundant features. This may result in a low-performance and extremely complex global model. 
Specifically, we can divide all features into qualified important features and negatively influential features, \emph{e.g.}, noisy features or redundant features, by their effects to the objective of the global model.
A desired VFL framework should enable all participants to jointly train a simple global model with a small number of important features, while eliminating negatively influential features.
The goal of feature selection in VFL is to simultaneously select a subset of features, and construct a global model $\hat{\bm{\theta}}$ with the objective by minimizing the risk,
\begin{equation}
\label{eq:vfl-feature-problem}
\begin{split}
    R(\bm{\theta},s)&:=\mathbb{E}_{X, Y} L(h(\theta_0, h_{n,1}, \cdots, h_{n,M}); y_n)\\
    &{\rm with} \,\,  h_{n,m}:=h_m(\theta_m; x_{n,m}\odot s_m), m\in[M],
\end{split}
\end{equation}
% \begin{equation}
% \label{eq:vfl-feature-problem}
% \begin{split}
%     R(\theta_0, \bm{\theta},\mathcal{S}, \mathcal{Q})&:=\mathbb{E}_{x_n, y_n} L(h(\alpha_0, u_{n,1}, \cdots, u_{n,M}); y_n)\\
%     &{\rm with} \,\,  u_{n,m}:=w_m\odot (h_{n,m}\odot q_m) \\
%     &h_{n,m}:=h_m(\theta_m; x_{n,m}\odot s_m), m\in[M],
% \end{split}
% \end{equation}
where $s_m=\{0, 1\}^{d_m}$ is the vector of indicator variables, and $s_{m,i}, i\in[d_m]$ are Bernoulli variables which indicate whether or not the $i$-th feature of client $m$ is selected. 
% That is, $s_{m,i}=1$ if $i\in \mathcal{S}_m$; otherwise, $s_{m,i}=0$.
We assume that all participants are semi-honest. They follow the exact protocol of VFL and feature selection, but are curious about others' private information.

% Here, we consider two phases of feature selection, which are 1) local clients conduct feature selection to remove their own irrelevant nuisance features or redundant features, and 2) the server conducts feature selection of top embedding vectors to remove redundant features among clients. Two-phase feature selection is necessary for reasons that, 1) The second phase is necessary for the reason that, due to the requirements of privacy preserving in VFL, any client has no knowledge of which local features are used by other clients. This would result in a number of similar or redundant features transmitted to the server, and participating in top model training, which would cause unnecessary computation and communication overhead. By selecting important features of cut layer, the server can inform clients to upload the selected ones to save computation and communication overhead.

\section{The Proposed \textnormal{FedSDG-FS} Approach}
\label{sec:methodoloy}
In this section, we first present the system architecture of FedSDG-FS. Then, we illustrate the key technique to enable feature selection to be performed jointly with model training under VFL settings. Finally, we present the details of the FedSDG-FS algorithms. 
% present the design of the proposed FedSDG-FS approach. We first present how to securely and efficiently initialize feature importance using Gini scores and PHE before the commencement of training (Algorithm \ref{alg:HE-feature1}). Then, we present the important feature selection method including the forward propagation for secure feature selection (Algorithm \ref{alg:Forward_VFL}), and the backward propagation for secure feature selection (Algorithm \ref{alg:Backward_VFL}) to achieve feature selection and VFL model training simultaneously.
\begin{figure}[t!]
	\begin{center}
		\includegraphics[width=0.98\linewidth,clip]{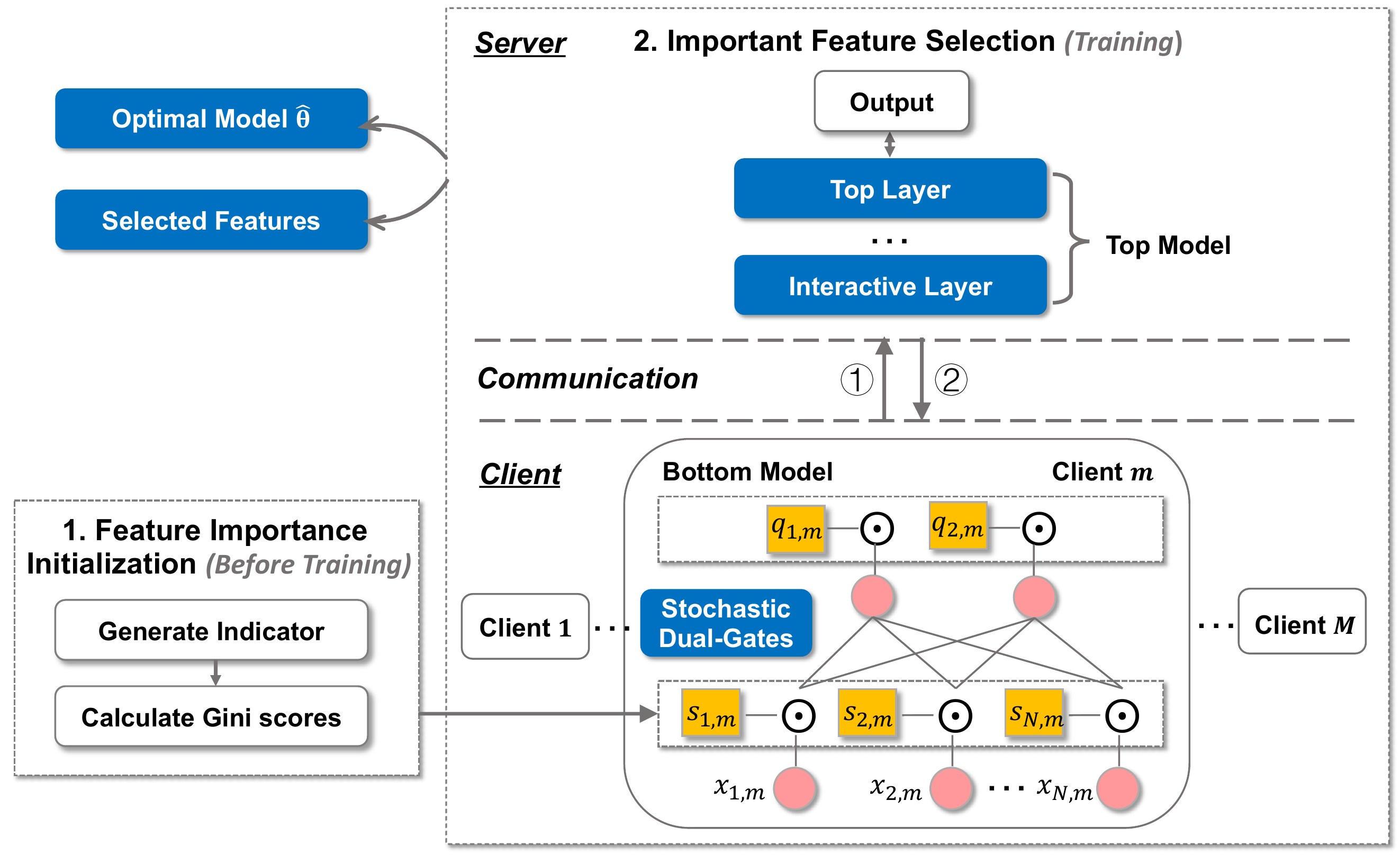}
 		\caption{System overview of FedSDG-FS. $\textcircled{1}$ Send encrypted embeddings, $\textcircled{2}$ send encrypted gradients.}
		\vspace{-0.2in}
		\label{fig:system_overview}
	\end{center}
	%\vspace{-0.05in}
\end{figure}
\subsection{System Overview}

FedSDG-FS consists of two modules (as shown in Fig. \ref{fig:system_overview}):

\textbf{1) Feature Importance Initialization before Training.}
To save feature selection costs, local clients first securely initialize feature importance based on Gini impurity and PHE, in cooperation with the server prior to the model training.
% Specifically, we propose a protocol based on Gini score and PHE to achieve secure noisy feature identification. 

\textbf{2) Important Feature Selection during Training.} 
After feature importance initialization, the server coordinates clients to select important features, while training the VFL model for improved performance.
Specifically, to fulfil the requirement that neither data nor labels can be obtained or inferred by any other party other than their original owners, 
we propose a secure FS approach which  includes forward propagation for secure feature selection, and backward propagation for secure feature selection, based on the proposed stochastic dual-gate, PHE and the randomized noise mechanism. In this way, FedSDG-FS determines the selected features and produces an optimal global model $\hat{\theta}$ with higher accuracy and fast convergence.

% A natural objective derived from derived from Eq. \eqref{eq:vfl-feature-problem} is the constrained empirical risk minimization. In VFL, local embedding vectors are transferred to the server, where the size of embedding vectors affects the communication cost. 

\subsection{Stochastic Dual-Gates for VFL}
\label{sec:Dual-gates}
To achieve accurate feature selection while simultaneously training the global model in VFL, we need to dynamically quantify the influence of features on the global model during training, and increase the probability of selection for highly influential features. In VFL, local embedding vectors are transferred to the server, where the size of embedding vectors affects the communication cost. To reduce communication overhead in feature selection, we first introduce stochastic dual-gates for VFL to efficiently approximate the probabilities of features and embedding vectors being selected. 
We re-express Eq. \eqref{eq:vfl-feature-problem} into minimizing the $l_0$ constrained risk:
\begin{equation}
\label{eq:vfl-L0}
\begin{split}
    R(\bm{\theta},s,q):=\mathbb{E}_{X,Y} &L(h(\theta_0, g_{n,1}, \cdots, g_{n,M}); y_n)\\
    &+\lambda \sum_{m} \big(|s_m|_0+|q_m|_0\big)\\
\end{split}
\end{equation}
where $h_{n,m}:=h_m(\theta_m; x_{n,m}\odot s_m)$, $ g_{n,m}=h_{n,m}\odot q_m$, $q_m=\{0, 1\}^{\underline{d_m}}$ is the vector of indicator variables, where $q_{m,i}, i\in[d_m]$ are Bernoulli variables and indicate whether or not the $i$-th dimension of embedding $h_{n,m}$ is selected for global model training. $\lambda$ is a weighting factor for the regularization. 
The $l_0$ norm penalizes the number of non-zero entries in the vectors $s_m$, $q_m$, thus encourages sparsity in the final estimates. Notice that $l_0$ norm induces no shrinkage on the actual values of the parameters, which is in contrast to $l_1$ regularization \cite{hans2009bayesian}.  

\begin{figure}[t!]
		\centering
		 \subfigure[Ratio of the same selected features by stochastic gate and Gini impurity ]{\label{fig:gini_initial_observe1}
    		\includegraphics[width = 0.46\linewidth, height=1.18in, trim=4 4 4 4]{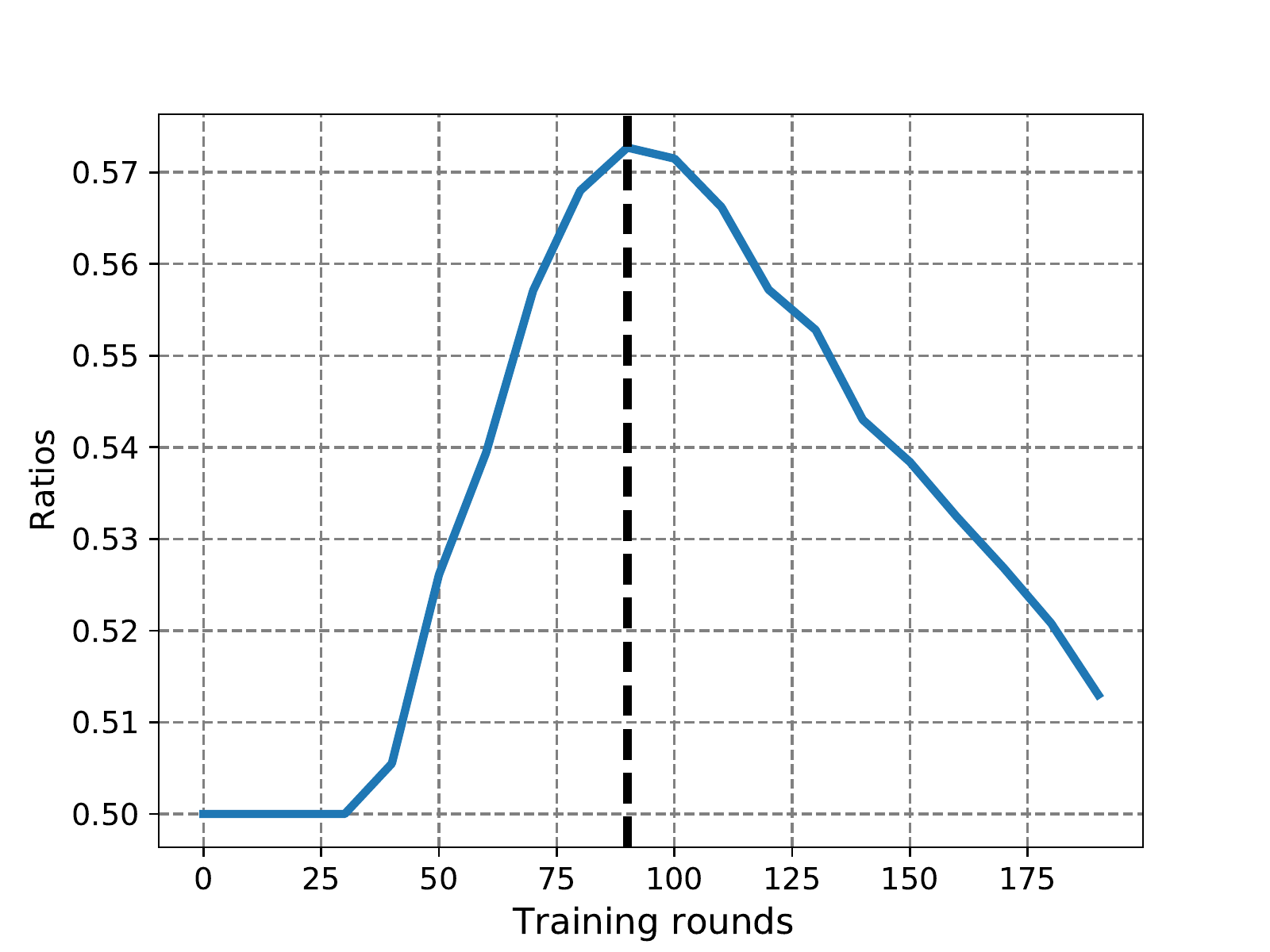}}
        	\subfigure[Comparison of the original gate method and FedSDG-FS ]{\label{fig:gini_initial_observe2}
        	\includegraphics[width = 0.463\linewidth, height=1.2in, trim=4 4 4 4]{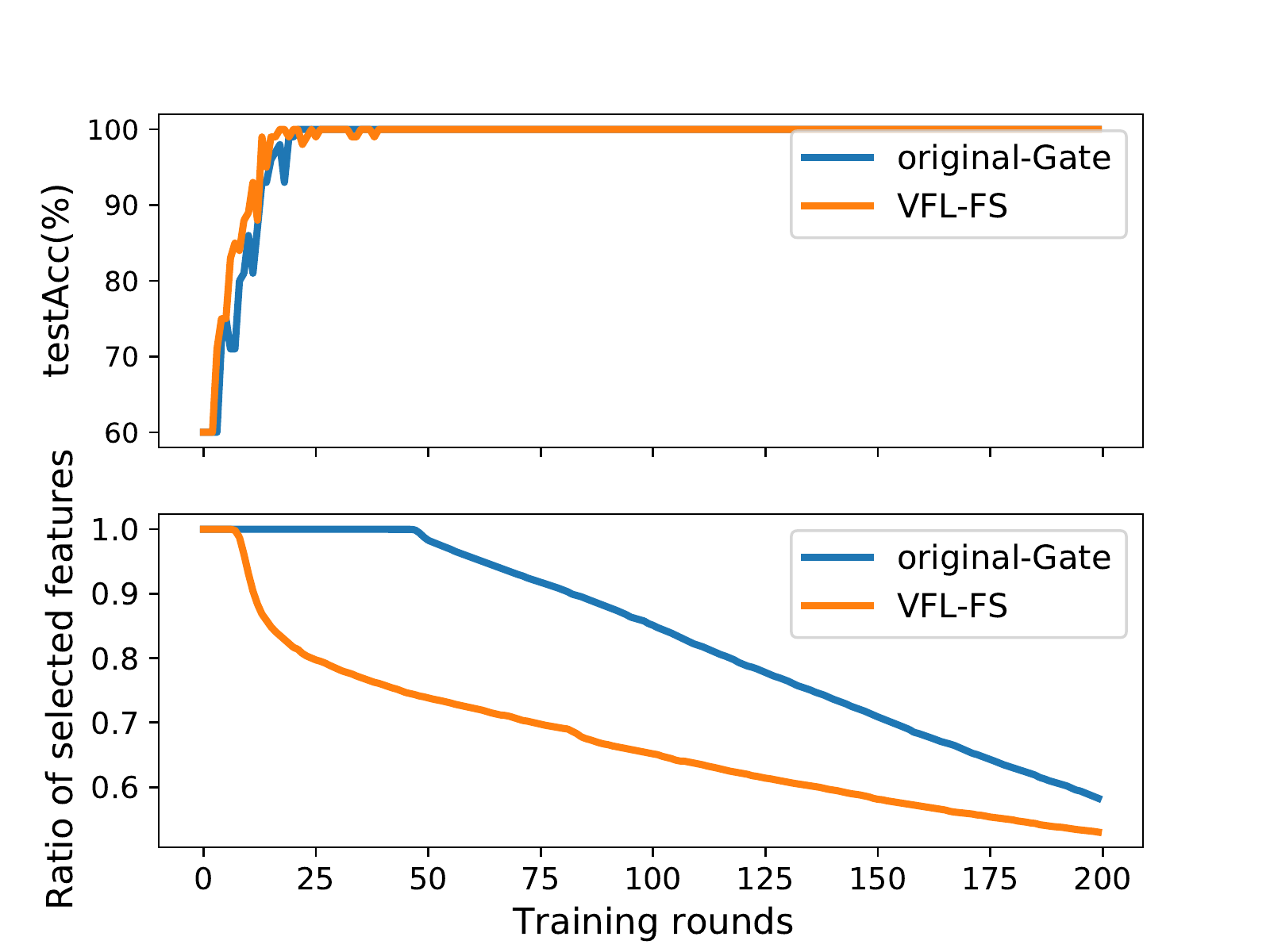}}
            	\vspace{-0.08in}
        	\caption{Example motivation of FedSDG-FS design. The vertical neural network is trained on 
        	the dataset \texttt{MADELON}.}	\label{fig:design_motivation}
\vspace{-0.1in}
\end{figure}

However, as the optimization of hard feature selection with binary masks suffers from high variance, we propose a secure Gaussian-based continuous relaxation for the Bernoulli variables for VFL. We approximate each element of $s_m, q_m$ to clipped Gaussian random variables parameterized by $\mu_m, \omega_m$ as $s_{m,i}=\max(0, \min (1, \mu_{m,i} + \rho_{m,i}))$, $q_{m,j}=\max(0, \min (1, \omega_{m,j} +\gamma_{m,j}))$, where $\rho_{m,i}, \gamma_{m,j}$ are drawn from $\mathcal{N}(0, \sigma^2)$, and $\mu_{m,i}, \omega_{m,j}$ can be learned during VFL training. 
Under the continuous relaxation, the regularization term in Eq. \eqref{eq:vfl-L0} is simply the sum of the probabilities that $\sum_{i\in [d_m]}P(s_{m,i}>0)+\sum_{j\in [\underline{d_m}]}P(q_{m,j}>0)$, and can be calculated by $\sum_{i\in [d_m]} \Phi (\frac{\mu_{m,i}}{\sigma})+\sum_{j\in [\underline{d_m}]} \Phi (\frac{\omega_{m,j}}{\sigma})$, where $\Phi(\cdot)$ is the cumulative distribution function (CDF) of the standard Gaussian distribution. 
By employing the continuous distribution, we can thus transform Eq. \eqref{eq:vfl-L0} into the following:
\begin{equation}
\label{eq:vfl-L0-relaxation}
\begin{split}
    R(\bm{\theta},\mu,\omega):=&\mathbb{E}_{X, Y} L(h(\theta_0, g_{n,1}, \cdots, g_{n,M}); y_n)\\
    &+\lambda \bigg(\sum_{m, i} \Phi \left(\frac{\mu_{m,i}}{\sigma}\right)+\sum_{m,j} \Phi \left(\frac{\omega_{m,j}}{\sigma}\right)\bigg).
\end{split}
\end{equation}

To optimize the objective of Eq. \eqref{eq:vfl-L0-relaxation}, we first differentiate it with respect to $\mu_m, \omega_m$.  
However, since the loss $L$ of the global model is calculated and stored at the server $S$, client $m$ performs the differentiation  using chain rules \cite{miller2017reducing} based on the Monte Carlo sampling gradient estimator, \emph{e.g.}, for $\mu_m$: 
\begin{equation}
\label{eq:monte-carlo}
    \frac{1}{C} \sum_{i\in [C]}
    \left[\frac{\partial L_n}{\partial g_{n,m} } \cdot \frac{\partial g_{n,m}}{\partial s_m } \cdot \frac{\partial s_{m,i}}{\partial \mu_{m,i} }
    \right] +\lambda \frac{\partial }{\partial \mu_m}\Phi \left(\frac{\mu_m}{\sigma}\right)
\end{equation}
where $C$ is the number of Monte Carlo samples. The calculation of gradient of estimator for $\omega_m$ is similar to Eq. \eqref{eq:monte-carlo}.
Thus, we can update $\mu_m, \omega_m$ via stochastic gradient descent.

Updating the parameters and conducting the above operations require access to all local training samples or training process, which are, however, obfuscated from any third party including the server. In addition, directly applying the stochastic gates to the clients' inputs would require a large number of parameters to be trained (\emph{e.g.}, $\mu_m$, $m\in [M]$), which slows down the convergence of the global model, and incurs significant computation and communication overhead, especially for high-dimension features. 

To address this challenge, we propose an efficient and secure feature selection framework, FedSDG-FS, which leverages Gini impurity for VFL to initialize the importance of individual features to facilitate feature selection. Then important features and significant local embeddings can be selected by the proposed stochastic dual gates, enhanced with PHE and the randomized noisy mechanism for privacy preservation.
To illustrate the motivation of the importance initialization, we make the following empirical observations.
Firstly, as illustrated in Fig. \ref{fig:gini_initial_observe1}, there can be a large ratio of the same features being selected by the Gini impurity \cite{li2021privacy} and by the stochastic gates in some training rounds. 
% First, as shown in the figure, a large proportion of the same features can be selected by both random gate training-based feature selection methods and gini-based feature selection methods
Secondly, Gini impurity initialization can speed up feature selection (Fig. \ref{fig:gini_initial_observe2}).
Moreover, the reason that Gini impurity cannot be directly used for feature selection is that it cannot take into account the specific VFL models and has no prior knowledge of the number of important features to select.
The feature importance initialization step can be accomplished by FedSDG-FS through two parameter transmissions with two encryption/decryption operations on the server based on Gini impurity and PHE, which significantly improves efficiency and privacy preservation. In this way, we can achieve efficient and secure feature selection as well as construct the global VFL model with high inference accuracy and fast convergence. 

\subsection{Feature Importance Initialization}
\label{sec:filter-gini}

\begin{algorithm}[t!]
 \SetAlgoVlined
%\small{
    \caption{Feature Importance Initialization for VFL}
    \label{alg:HE-feature1}
    \SetKwInOut{Input}{Input}\SetKwInOut{Output}{Output}
%    \begin{algorithmic}[1]
    \Input{Server $S$,
    clients $m$ }
    \Output{Initialized feature importance}
    {
    {\bf Server $S$}\\    \quad Generate an indicator matrix $A$, $[\![A]\!]\leftarrow Enc(A)$\\ \quad Send $[\![A]\!]$ to all clients\\
    }
  {
  {\bf Client $m$}\\
    \quad Induce a partition $U_{m,1}\cup U_{m,2}\cup\cdots\cup U_{m,b}$ of $U_m$\\
   \quad Calculate $[\![p_{m,k}]\!] \leftarrow \sum_{a\in I(U_{m,i})}[\![A]\!]_{a,k}/|U_{m,i}|$\\
\quad Calculate $[\![p_{m,k}]\!]^2$ with the protocol in  \cite{erkin2009privacy}\\
    \quad $[\![G(U_{m,i})]\!]\leftarrow 1-\sum_{k\in [c]} [\![p_{m,k}]\!]^2$\\ 
    \quad $[\![G(f_{m,j})]\!]\leftarrow \sum_{i=1}^c \frac{|U_{m,i}|}{|U_m|}\cdot [\![G(U_{m,i})]\!]$\\ 
    \quad Send $[\![G(f_{m,j})]\!], j\in[d_m]$ to the server\\
    }
    {
    {\bf Server} S\\
    % \quad Initializes an index set $Q_m\leftarrow \varnothing$ for client $m$ \\
    \quad $G(f_{m,j})\leftarrow Dec([\![G(f_{m,j})]\!])$\\ 
    \quad Send $G(f_{m,j}), j\in [d_m]$ to client $m$\\
    }
    {
    {\bf Client $m$}\\
    \quad Initialize $\mu_{m,j}\propto \frac{1}{G(f_{m,j})}
    $ \\
    \textbf{Return} feature importance initialization $\mu_{m,j}, j\in [d_m]$
    }
   
\end{algorithm}
% When a VFL task arrives, the server and local clients securely and collaboratively filters noisy features with the Gini impurity before VFL training. 
$M$ clients have a set $U=\{U_1, \cdots, U_M\}$ of $N$ samples, and the corresponding $c$ class labels are stored at the server. 
% For $M$ clients, we have a set $U=\{U_1, \cdots, U_M\}$ of $N$ samples and the corresponding $c$ class labels stored at the server.
For client $m$, if the $j$-th feature $f_{m,j}$ is a discrete feature that can assume $b$ values, then it induces a partition $U_{m,1}\cup \cdots \cup U_{m,b}$ of the set $U_m$ in which $U_{m,i}$ is the set of instances with the $i$-th value for $f_{m,j}$. The Gini impurity of $U_{m,i}$ is defined as $G(U_{m,i}) = 1-\sum_{k\in [c]} p_{m,k}^2$,
% \begin{equation}
% \label{eq:gini-imp}
%     G(U_{m,i})=\sum_{k\in [c]} p_{m,k} \cdot (1-p_{m,k}) = 1-\sum_{k\in [c]} p_{m,k}^2
% \end{equation}
where $p_{m,k}$ is the probability of a randomly selected instance from $U_{m,i}$ belonging to the $k$-th class. 
The Gini score of feature $f_{m,j}$ is calculated as $G(f_{m,j})=\sum_{i\in [b]} \frac{|U_{m,i}|}{|U_m|}\cdot G(U_{m,i})$, 
% \begin{equation}
% \label{eq:gini-score}
%     G(f_{m,j})=\sum_{i\in [c]} \frac{|U_{m,i}|}{|U_m|}\cdot G(U_{m,i})
% \end{equation}
where $G(f_{m,j})$ measures the likelihood of a randomly selected instance being misclassified.
% , and the features with the lowest Gini scores are retained. 
If $f_{m,j}$ is a feature with continuous values, then $G(f_{m,j})$ is defined as the weighted average of the Gini impurities of a set of discrete feature values. We use the Paillier as the PHE method which supports homomorphic addition of two ciphertexts and homomorphic multiplication between a plaintext and a ciphertext. The calculation of $p_{m,k}$ requires collaboration between client $m$ and the server. Thus, we design an efficient and secure collaborative calculation protocol. 
% to protect the data of all clients and labels of the server. 

% Before VFL training, the server and clients collaborate to filter noisy features based Gini scores and homomorphic encryption. 
Specifically, the server first generates an indicator matrix $A$ with a size of $N\times c$, where $A_{n,k}=1$ indicates the category of the $n$-th sample is $k$; otherwise, $A_{n,k}=0$. 
% whether the category of the $n$-th sample is $k$ for $n\in [N]$, $k\in [c]$. That is, if the category of the $n$-th sample is $k$, then $A_{n,k}=1$; otherwise, $A_{n,k}=0$. 
Then, the probability $p_{m,k}$ can be calculated as $p_{m,k}=\sum_{a\in I(U_{m,i})}A_{a,k}/|U_{m,i}|$ for client $m$, where $I(U_{m,i})$ denotes the index set of instances from $U_{m,i}$. 
To prevent the private label information from being leaked, the server encrypts the matrix $A$, and sends $[\![A]\!]$ to all clients. 
Then, client $m$ calculates the probability $[\![p_{m,k}]\!]=\sum_{a\in I(U_{m,i})}[\![A]\!]_{a,k}/|U_{m,i}|$ and uses the protocol in  \cite{erkin2009privacy} to compute the square of $[\![p_{m,k}]\!]$ as follows. 
Firstly, client $m$ generates a random value $r$ and computes $[\![u_{m,k}]\!] = [\![p_{m,k}+r]\!]$, such that $p_{m,k}^2$ equals $u_{m,k}^2 - 2u_{m,k}\cdot r + r^2$ and $[\![- 2u_{m,k}\cdot r + r^2]\!]$ can be locally computed by the client.
Then, client $m$ sends $[\![u_{m,k}]\!]$ to the server. The server decrypts it, computes and sends $[\![u_{m,k}^2]\!]$ to client $m$.
Finally, the client computes $[\![p^2]\!] = [\![u_{m,k}^2 - 2u_{m,k}\cdot r + r^2]\!]$. 
After calculating $[\![p^2]\!]$, client $m$ calculates the Gini impurity $[\![G(f_{m,j})]\!]$ of feature $f_{m,j}$, and sends them to the server.
The server then decrypts them, and assigns larger initial importance values to features with smaller Gini values. During this process, only the server learns the Gini scores of client $m$'s features, while other parties learn nothing.
The main steps are shown in Algorithm \ref{alg:HE-feature1}.
% Then the server decrypts them and identifies those indices of features whose Gini scores are greater than $\tau$, and then informs clients to filter out features with corresponding indices. If the number of remaining features of a client is less than $\beta$, the client will not be allowed to participate in VFL training.  After this computation, only the server learns Gini scores of  client $m$'s features, while other parties learn nothing.
% The total communication cost is $O(n'|\mathcal{Y}|)$, while the existing method requires $O(n'(|\mathcal{Y}|+\sum_{k=1}^{N'}(d_k+1))+\sum_{k=1}^{N'} d_k^2)$ operations \cite{pansecure}. 
% Now the server can solve the optimization problem (Eq. \ref{eq:selection_goal}) by greedily choosing the clients with the largest $\sum_{i=1}^{d_k} I[G(F_{k,j})>\tau]/b_k$, $k\in[N']$ until the budget $B$ runs out. 

 \subsection{Secure Important Feature Selection}

%  To achieve accurate feature selection as well as high-performance model training, we design a secure and importance-aware feature selection approach for VFL, which leverages secure forward propagation and secure backward propagation methods to achieve secure feature selection and VFL training. 
 
%  In addition, to achieve efficient feature selection, we design a importance sampling based method to accelerate the approximation speed and model convergence speed.  
%  We note the homomorphic encryption as $[\![\cdot]\!]$. We detail the privacy issue and the algorithm in the following section. 
    
\begin{algorithm}[b!]
 \SetAlgoVlined
    \caption{Forward Propagation for Secure Feature Selection}
    \label{alg:Forward_VFL}
    \SetKwInOut{Input}{Input}\SetKwInOut{Output}{Output}
    \Input{$M$ clients with $N$ samples $\{x_n, y_n\}_{n=1}^N$, $x_{n,m}\in \mathbb{R}^{d_m}$}
    \Output{Global model $\hat{\theta}$, indicator vector $\{s_m\}_{m=1}^M$}
    {
    Initialize model $\theta_0:=\{\alpha_0, w_1, w_2, \cdots, w_M\}$,  $\{\theta_i\}_{i=1}^M$, noise $\epsilon_{acc}$; initialize
    $\mu_{m}$ with Algorithm \ref{alg:HE-feature1}, $\omega_m\in \mathbb{R}^{\underline{d_m}}$\\
    }
    {\bf Client $m$, $m\in [M]$ \\}
    {
    \quad Select datum (or data mini-batch) $x_{n,m}$\\
    \quad Sample $\rho_{m,i}, \gamma_{m,j}\sim \mathcal{N}(0, \sigma^2)$, $i\in [d_m], j\in [\underline{d_m}]$\\
    \quad  Compute $s_{m,i}=\max (0, \min (1, \mu_{m,i}+\rho_{m,i}))$ \\
    \quad  $q_{m,j}=\max (0, \min (1, \omega_{m,j}+\gamma_{m,j}))$ \\
    \quad  $R_m=\sum_{i\in [d_m]} \Phi \big(\frac{\mu_{m,i}}{\sigma}\big)+\sum_{j\in [\underline{d_m}]} \Phi \big(\frac{\omega_{m,j}}{\sigma}\big)$\\
    \quad $h_{n,m}\leftarrow h_m(\theta_m; x_{n,m}\odot s_m), g_{n,m}=h_{n,m}\odot q_m$\\ \quad $[\![g_{n,m}]\!]\leftarrow Enc(g_{n,m})$, sends $[\![g_{n,m}]\!]$ to the server\\
    % }
    }
    {
    {\bf Server $S$} \\
    \quad Calculate the noisy weight $\widetilde{w}_m\leftarrow w_m+\epsilon_{acc}$, $m\in[M]$\\
    \quad Compute $[\![\widetilde{z}_{n,m}]\!]\leftarrow [\![g_{n,m}]\!] \cdot \widetilde{w}_m$\\
    \quad Add random noise $[\![\widetilde{z}_{n,m}+\epsilon_s]\!]\leftarrow  [\![\widetilde{z}_{n,m}]\!]+\epsilon_s$\\
    \quad Send $[\![\widetilde{z}_{n,m}+\epsilon_s]\!]$ to client $m$\\
    }
    {\bf Client $m$, $m\in [M]$ \\}
    {
    \quad $\widetilde{z}_{n,m}+\epsilon_s\leftarrow Dec([\![\widetilde{z}_{n,m}+\epsilon_s]\!])$\\
    \quad Remove noise  $z_{n,m}+\epsilon_s\leftarrow \widetilde{z}_{n,m}+\epsilon_s -\epsilon_{acc} g_{n,m}$\\
    \quad Send $z_{n,m}+\epsilon_s$ to the server\\
    }
    {\bf Server $S$}\\
    {
    \quad Remove noise $z_{n,m}\leftarrow z_{n,m}+\epsilon_s-\epsilon_s$\\
    \quad Compute 
    $L_n \leftarrow L(h(\alpha_0, z_{n,1}, \cdots, z_{n,M}); y_n)$\\
    }
	\textbf{Return} the loss $L_n$
\end{algorithm}

% \subsubsection{\bf Qualified Client Identification.} 
% that is, after several rounds of global training, when we find that some local clients' models have converged, then we stop updating those models and setting cutoff values, \emph{e.g.}, 0.5, of $\mu_m$s. 

% As illustrated in Section \ref{sec:importance-highlight}, the server and clients can identify which clients should conduct feature selection and model training using the distance between two consecutive local models $||\theta_m^{t+1}-\theta_m^t||$ of client $m$. The clients locally calculates the distance $D_m=\frac{1}{a_t}\sum_{t=i}^{i+a_t}||\theta_m^{t+1}-\theta_m^t||$. If the distance of client $m$ is less than the threshold $\tau$, then he is a client. 

% We use the difference between two consecutive local models $||\theta_m^{t+1}-\theta_m^t||$ of client $m$ to characterize whether the local model converges. 

\subsubsection{\bf Forward Propagation on Clients} 
Client $m$ randomly selects a private datum (or mini-batch) $x_{n,m}$, and calculates the indicator $s_{m,i}$ for each feature $f_{m,i}$, $i\in [d_m]$. Then, it calculates the embedding vector $h_{n,m}$ using the local model $\theta_m$ and the masked embedding $g_{h,m}=h_{n,m}\odot q_m$, and encrypts it with PHE to obtain $[\![g_{n,m}]\!]=Enc(g_{n,m})$, which is sent to the server. 
% To prevent information leakage caused by plaintext transmission, client $m$ encrypts the embedding  $h_{n,m}$ with partially homomorphic encryption (PHE), \emph{e.g.}, Paillier, to $[\![h_{n,m}]\!]$, and sends it to the server. 

\subsubsection{\bf 
Forward Propagation on the Server} After receiving the encrypted embedding $[\![g_{n,m}]\!]$, the server calculates the weighted vector $[\![z_{n,m}]\!]=[\![g_{n,m}]\!]\odot w_m$, and performs the forward propagation of the top model.  
% After receiving the encrypted embedding  $[\![h_{n,m}]\!]$, the server calculates the weighted sum of it $[\![z_{n,m}]\!]=[\![h_{n,m}]\!]\odot w_m$ to conduct forward propagation. 
Since the non-linear activation function on the top model cannot be calculated on the encrypted data, the weighted vector $[\![z_{n,m}]\!]$ should be sent back to client $m$ for decryption. 
However, sending the weighted vector directly without any protection would leak the prediction to the client (\emph{e.g.}, client $m$ can use the activation prediction pair $(z_{n,m}, g_{n,m})$ to infer activation values and weights of the top model). 
To prevent this, the server adds random noises $\epsilon_s$ on $[\![z_{n,m}]\!]$, and sends $[\![z_{n,m}+\epsilon_s]\!]$ to client $m$. Then, client $m$ decrypts the noisy weighted sum $[\![z_{n,m}+\epsilon_s]\!]$, and sends $z_{n,m}+\epsilon_s$ to the server. Finally, the server removes the noise and computes the activation for the next layer. The process repeats until the final layer is reached. 
\begin{algorithm}[b!]
 \SetAlgoVlined
    \caption{Backward Propagation for Secure Feature Selection}
    \label{alg:Backward_VFL}
    \SetKwInOut{Input}{Input}\SetKwInOut{Output}{Output}
    \Input{Loss $L_n$ on the server, target $\{y_n\}_{n=1}^N$, learning rates $\eta_0$, $\eta_m$}
    \Output{Global model $\hat{\theta}$, indicator vector $s_m, q_m, m\in [M]$}
    % {
    % Initialize model parameters $\theta_0:=\{\alpha_0, w_1, w_2, \cdots, w_M\}$,  $\{\theta_i\}_{i=1}^M$\\
    % }
    {
    \textbf{Server $S$}\\ 
    \quad Compute the gradients $[\![\frac{\partial L_n}{\partial w_m}]\!]\leftarrow \frac{\partial L_n}{\partial z_{n,m}} \cdot [\![g_{n,m}]\!]$, $\frac{\partial \widetilde{L}_n}{\partial g_{n,m}}\leftarrow\frac{\partial L_n}{\partial z_{n,m}} \cdot \widetilde{w}_m$, $\frac{\partial L_n}{\partial \alpha_0}$
    \\
    \quad Add noise $[\![\frac{\partial L_n}{\partial w_m} +\epsilon_s]\!] \leftarrow [\![\frac{\partial L_n}{\partial w_m}]\!] +\epsilon_s$\\
    \quad Send $[\![\frac{\partial L_n}{\partial w_m}+\epsilon_s]\!]$ to client $m$\\
    }
    {
    {\bf Client $m\in [M]$}\\
    \quad $\frac{\partial L_n}{\partial w_m}+\epsilon_s\leftarrow Dec([\![\frac{\partial L_n}{\partial w_m}+\epsilon_s]\!])$\\
    \quad Add noise $\frac{\partial \widetilde{L}_n}{\partial w_m} +\epsilon_s\leftarrow \frac{\partial L_n}{\partial w_m} +\epsilon_s-\frac{\epsilon_m}{\eta_0}$\\
    \quad Encrypt noise $[\![\epsilon_{acc}]\!]\leftarrow Enc(\epsilon_{acc})$\\
    \quad Accumulate noise $\epsilon_{acc}\leftarrow \epsilon_{acc}+\epsilon_m$\\
    \quad Send $\frac{\partial \widetilde{L}_n}{\partial w_m}+\epsilon_s$, and $[\![\epsilon_{acc}]\!]$ to the server\\
    }
    {
    {\bf Server $S$}\\
    \quad Remove noise $\frac{\partial \widetilde{L}_n}{\partial w_m}\leftarrow \frac{\partial \widetilde{L}_n}{\partial w_m} +\epsilon_s-\epsilon_s$\\
    \quad Update  $\theta_0=\{w_1,\cdots, w_m,\alpha_0\}$:   $\widetilde{w}_m\leftarrow \widetilde{w}_m-\eta_0 \frac{\partial \widetilde{L}_n}{\partial w_m}$, $\alpha_0\leftarrow \alpha_0-\eta_0 \nabla_{\alpha_0}L_n$ \\
    \quad Remove noise $[\![\frac{\partial L_n}{\partial g_{n,m}}]\!] \leftarrow \frac{\partial \widetilde{L}_n}{\partial g_{n,m}}  - [\![\epsilon_{acc}]\!]\cdot \frac{\partial L_n}{\partial z_{n,m}} $  \\
    \quad Send $[\![\frac{\partial L_n}{\partial g_{n,m}}]\!]$ to client $m$\\
    }
    {
    {\bf Client $m\in [M]$}\\
    \quad $\frac{\partial L_n}{\partial g_{n,m}} \leftarrow [Dec([\![\frac{\partial L_n}{\partial g_{n,m}}]\!])$\\ 
    \quad Calculate $\frac{\partial L_n}{\partial \mu_{m}}, \frac{\partial L_n}{\partial \omega_{m}}, \frac{\partial L_n}{\partial \theta_m} $\\
    % \quad $\frac{\partial L_n}{\partial \theta_m}\leftarrow \frac{\partial L_n}{\partial g_{n,m}} \cdot \frac{\partial g_{n,m}}{\partial \theta_m}$\\
    \quad Update $\mu_m\leftarrow \mu_m-\eta_m \big (\frac{\partial L_n}{\partial \mu_m} +\lambda \frac{\partial R_m}{\partial \mu_m} \big)$\\
    \quad $\omega_m\leftarrow \omega_m-\eta_m \big (\frac{\partial L_n}{\partial \mu_m} +\lambda \frac{\partial R_m}{\partial \omega_m} \big)$, $\theta_m\leftarrow \theta_m-\eta_m \frac{\partial L_n}{\partial \theta_m}$\\
    }
	\textbf{Return} the global model ${\theta}=\{\theta_m\}_{m=0}^M $.
\end{algorithm}

Another problem is that the server holds both $w_m$ and $z_{n,m}$, and can easily infer $h_{n,m}$ via linear regression. To avoid this, the server should use the noisy weight $\widetilde{w}_m$ to calculate the weighted vector $[\![\widetilde{z}_{n,m}]\!]\leftarrow [\![g_{n,m}]\!] \odot \widetilde{w}_m$, where $\widetilde{w}_m=w_m+\epsilon_{acc}$, $\epsilon_{acc}$ is generated by the client. 
% We will discuss how to add $\epsilon_{acc}$ to $w_m$ in Section \ref{sec:BP-server}. 
The forward propagation for secure feature selection is shown in Algorithm \ref{alg:Forward_VFL}.

\subsubsection{\bf Backward Propagation on the Server}
% \subsubsection{Secure Backward Propagation on the Server}
\label{sec:BP-server}
To update the global model, two gradients need to be computed first, the loss gradients w.r.t. the weight of the interactive layer $\frac{\partial L_n}{\partial w_m}$, and the embedding vector $\frac{\partial L_n}{\partial g_{n,m}}$.
Since these two gradients are linear transformations of either $g_{n,m}$ or $w_m$, both the server and client $m$ can derive what they want to acquire via regression. 
To this end, we design the following secure backward propagation method. 

Specifically, the server first calculates the following gradients: $[\![\frac{\partial L_n}{\partial w_m}]\!],  \frac{\partial \widetilde{L}_n}{\partial g_{n,m}}, \frac{\partial L_n}{\partial \alpha_0}$. 
If the server updates $[\![w_m]\!]$ by $[\![w_m]\!]=w_m-\eta_m [\![\frac{\partial L_n}{\partial w_m}]\!]$, this would result in two encrypted quantities in calculating the weighted vector $z_{n,m}=[\![w_m]\!] [\![g_{n,m}]\!]$, which is incompatible with PHE. 
To avoid this, the server needs to send $[\![\frac{\partial L_n}{\partial w_m}]\!]$ to client $m$, and receive the decrypted gradient $\frac{\partial L_n}{\partial w_m}$ back. 
However, sending $[\![\frac{\partial L_n}{\partial w_m}]\!]$ directly to client $m$ would leak information about both parties, because the server holds $\frac{\partial L_n}{\partial z_{n,m}}$ and client $m$ holds $h_{n,m}$.
Thus, both the server and client $m$ need to add random noises to the encrypted gradient of weights $\frac{\partial L_n}{\partial z_{n,m}}$ before sending them to the other party, and update the parameters (see lines 4-10 of  Algorithm \ref{alg:Backward_VFL}). 
Note that the noise $\epsilon_s$ generated by the server can be removed when the gradient $\frac{\partial \widetilde{L}_n}{\partial w_m}$ still contains noise, where $\frac{\partial \widetilde{L}_n}{\partial w_m}=\frac{\partial L_n}{\partial w_m}-\frac{\epsilon_m}{\eta_0}$.  
With $\frac{\partial \widetilde{L}_n}{\partial w_m}$, the server updates the weights as  $\widetilde{w}_m^{t+1}=w_m^t-\eta(\frac{\partial L_n}{\partial w_m} -\frac{\epsilon_m}{\eta_0}) =w_m^{t+1}+\epsilon_m$.
% \begin{equation}
% \begin{split}
%     \widetilde{w}_m^{t+1}=w_m^t-\eta(\frac{\partial L_n}{\partial w_m}
%     -\frac{\epsilon_m}{\eta})
%     =w_m^t-\eta \frac{\partial L_n}{\partial w_m}+\epsilon_m =w_m^{t+1}+\epsilon_m
% \end{split}
% \end{equation}

It can be observed that the noise $\epsilon_m$ will accumulate in weights $w_m$ in each iteration.
If we take the accumulated noise as $\epsilon_{acc}=\sum_{m=1}^M \sum_{i=1}^t \epsilon_m^i$, the true weights used in forward and backward propagation should be $w_m^{t+1}=\widetilde{w}_m^{t+1}-\epsilon_{acc}$. To perform the correct forward operation, client $m$ needs to remove the noise by subtracting $g_{n,m}\epsilon_{acc}$ from the noisy weighted vector $\widetilde{z}_{n,m}$. Similarly, the extra noise should be added to $\frac{\partial \widetilde{L}_n}{\partial g_{n,m}}$, and removed before backpropagation by client $m$. 
To achieve this, client $m$ needs to send the encrypted noise $[\![\epsilon_{acc}]\!]$ to the server, and the server calculates the true gradient via $[\![\frac{\partial L_n}{\partial g_{n,m}}]\!] = \frac{\partial \widetilde{L}_n}{\partial g_{n,m}}  - [\![\epsilon_{acc}]\!]\cdot \frac{\partial L_n}{\partial z_{n,m}}$, and sends the encrypted gradient $[\![\frac{\partial L_n}{\partial g_{n,m}}]\!]$ to the client $m$.
\subsubsection{\bf Backward Propagation on Clients}
% \subsubsection{Secure Forward Propagation on Clients}
The client $m$ first decrypts the gradient $[\![\frac{\partial L_n}{\partial g_{n,m}}]\!]$ received from the server. Then,  it updates the local model $\theta_m$ and the variable $\mu_m, \omega_m$. In this way, model update and feature selection can be accomplished simultaneously. 
The entire secure backpropagation approach is detailed in Algorithm \ref{alg:Backward_VFL}.

\subsection{Convergence Analysis}
We present convergence results for FedSDG-FS through two steps. First, we show that there is an equivalence between our proposed $l_0$ constrained optimization for feature selection and optimization over Bernoulli distribution  through Mutual Information (MI). Then, we present the convergence results of the gradient decent  methods for optimizing the $l_0$ constrained optimization.  
Without loss of generality, we only consider the bottom level stochastic gates here.
The goal of feature selection is to find the subset of features $Q$ that has the highest MI with the target variable $Y$. We can then formulate the task as selecting $Q$ such that the MI $I(\cdot)$ between $X_Q$ and $Y$ is maximized:
\begin{equation}
    \max_{Q} I(X_{Q}, Y)  \quad s.t. \quad |Q| = k.
\end{equation}
Then, under the mild assumption that there exists an optimal subset of indices $Q^*$, the equation above is equivalent to
\begin{equation}
    \max_{\mathbf{0} \leq \boldsymbol{\pi} \leq \mathbf{1}} I(X \odot \tilde{Q} ; Y) \quad s.t.  \quad \sum_{i} \mathbb{E}\left[\tilde{Q}_{i}\right] \leq k, 
\end{equation}
where $\tilde{Q}$ are independently sampled from the Bernoulli distribution with parameter ${\bm{\pi}}$. Then, we can rewrite this constrained optimization problem as a penalty optimization problem, which is the same as Eq. \eqref{eq:vfl-L0}:
\begin{equation}
    R = \min L(X \odot \tilde{Q}; Y) + \lambda |\tilde{Q}|
\end{equation}
So far, we have proved the equivalence between the proposed $l_0$ constrained optimization and the selection of the optimal feature subset. Next, we give the convergence results of the $l_0$ constrained optimization for feature selection. 

\textbf{Assumption 1.} 
\emph{The gradient $\frac{\partial R( \theta)}{\partial \theta_0}$ is $K$-Lipschitz continuous, and $\frac{\partial R(\bm \theta)}{\partial \theta_m}, m\in [M]$ is $K_m$-Lipschitz continuous. } 

\textbf{Theorem 1.} Under Assumption 1, and the assumption that $R(\theta)$ is $\rho$-strongly convex, if $\eta^t=\frac{1}{\rho \min_{m}(t+T_0)}$ with the constant $T_0>0$. Then the convergence rate is  $\mathcal{O}(1/T)$.

\textbf{Time and storage complexity analysis.} The time complexity of the algorithm is $\mathcal{O}(\frac{K ||\theta_0-\hat{\theta}||^2}{2\epsilon }), \epsilon=\frac{1}{2(\eta^t+\eta_m^t)T} (||\theta_0-\hat{\theta}||^2)$. The storage complexity is $\mathcal{O}(|\theta|+|s|+|q|)$, where $|\cdot|$ denotes the parameter size, and the communication cost is $\mathcal{O}\left(|q|\frac{K||\theta_0-\hat{\theta}||^2}{\epsilon}\right)$.
% for only the embedding vectors are transmitted.

\section{Experimental Evaluation}
\label{sec:exp}

\subsection{Experiment Configuration}
% \subsubsection{\textbf{Datasets}} 
\textbf{1) Datasets.}
We use 9 datasets with 4 types of data: tabular data, images, texts and audios. These include 2 synthetic datasets, \texttt{MADELON} \cite{guyon2004result} and \texttt{FRIEDMAN} \cite{friedman1991multivariate}; and 7 real-world datasets, \texttt{ARCENE} \cite{ARCENE}, \texttt{BASEHOCK} \cite{real-world-datasets}, \texttt{RELATHE} \cite{real-world-datasets}, \texttt{PCMAC} \cite{real-world-datasets}, \texttt{GISETTE} \cite{gisette}, \texttt{COIL20} \cite{coil20} and \texttt{ISOLET} \cite{isolet}. 
The synthetic datasets are derived from the feature selection challenge \cite{guyon2004result}, where \texttt{MADELON} consists of 5 informative features, 15 redundant features constructed by linear combinations of those 5 informative features, and 480 noisy features, while \texttt{FRIEDMAN} consists of 5 informative and 995 noisy features. 
For the real-world datasets, most of them are collected from the ASU feature selection database online \cite{real-world-datasets}. The descriptions of all datasets are listed in Table \ref{tab:dataset}.
We employ two clients in our settings, where we divide features into two parts randomly for every dataset, and assign each part to clients A and B. The labels are located in the server. For the text, image and audio datasets, we divide the features randomly by rows for the clients.

\begin{table}[ht!]
	\centering
% 	\vspace{-0.2in}
	\caption{Description of datasets for empirical evaluations}
	\vspace{-0.1in}
	\label{tab:dataset}
	\resizebox{1\linewidth}{!}
	{
		\begin{tabular}{l c  cc c l}
			\toprule \textbf{Dataset}&\textbf{ Features}&\textbf{Train size} & \textbf{Test size}& \textbf{Classes}& \textbf{Type}  \\
			 \hline 
			 \texttt{MADELON}
			 & 500 & 2,000 & 2,400 & 2 &  Tabular \\
			 \texttt{FRIEDMAN} & 1,000 & 750 & 250 & 2 &  Tabular \\
			 \texttt{ARCENE} & 10,000 & 1,400 & 600 & 2 & Tabular \\
			 \texttt{BASEHOCK} & 7,862 & 1,594 & 398 & 2 & Text \\
			 \texttt{RELATHE} & 4,322 & 2,320 & 2,088 & 2 & Text \\
			 \texttt{PCMAC} & 3,289 & 1,554 & 388 & 2 & Text \\
			 \texttt{GISETTE} & 5,000 & 5,600 &1,400 & 2 &  Image\\
			 \texttt{COIL20} & 1,024 & 1,008 & 432 & 20 &  Image\\
			 \texttt{ISOLET} & 617 &1,248 & 312 & 26 & Audio \\
			\toprule
		\end{tabular}
		\vspace{-0.1in}
	}
\end{table}

% \subsubsection{\textbf{VFL Models}}
\textbf{2) VFL Models.}
We have implemented the typical logistic regression model for VFL \cite{hardy2017private} on \texttt{FRIEDMAN}, and neural networks for VFL \cite{zhang2020additively} on the other 8 datasets (see Table \ref{tab:model}). 
We run VFL models until a pre-specified test accuracy is reached, or a maximum number of iterations has elapsed.
In addition, training the dual-gates until convergence may sometimes cause overfitting of the model, where we set the cutoff value of the variables and perform early stopping. 
We use the Paillier as the PHE method. We use the Adam optimizer, and set learning rate $\eta=0.03$, batch size $b=128$, weight factor $\lambda=0.1$. We test the accuracy of the global model on the hold-out test datasets.
We build our VFL models with Flower 0.19.0 \cite{flower-FL} and Pytorch 1.8.1 \cite{Pytorch}.
All the experiments are performed on Ubuntu 16 operating system equipped with a 12-core i7 Intel CPU, 64G of RAM and 4 Titan X GPUs. 

\begin{table}[ht!]
	\centering
	\vspace{-0.1in}
	\caption{Settings for training different VFL models.}
	\vspace{-0.1in}
	\label{tab:model}
	\resizebox{1\linewidth}{!}
	{
		\begin{tabular}{l| c| l}
			\toprule \textbf{Model}&\textbf{ \# of parameters}& \textbf{Task}\\
			 \hline 
			 \texttt{VFLNN-MADELON}
			 &130,552&Two-class classification \\
			 \hline \texttt{VFLLR-FRIEDMAN} &130,501&Regression\\
			 \hline \texttt{VFLNN-ARCENE} & 1,030,552 & Cancer detection
			 \\
			 \hline \texttt{VFLNN-BASEHOCK} &516,752&Text classification \\
			 \hline \texttt{VFLNN-RELATHE} &462,752 &Text classification \\
			 \hline \texttt{VFLNN-PCMAC} &359,452&Text classification  \\
			 \hline \texttt{VFLNN-GISETTE} &530,552&Digit number recognition\\
			 \hline \texttt{VFLNN-COIL20} &109,360&Face image recognition\\
			 \hline \texttt{VFLNN-ISOLET} &93,476&Letter-name recognition \\
			\toprule
		\end{tabular}
		\vspace{-0.1in}
	}
\end{table}
\begin{figure*}[t!]
	\begin{center}
		\begin{minipage}[c]{0.48\linewidth}
			\subfigure[\texttt{VFLNN-MADELON}]{\label{fig:madelon_gini_testAcc}
			    \includegraphics[width = 0.48\linewidth, height=1.16in, trim=4 4 4 4]{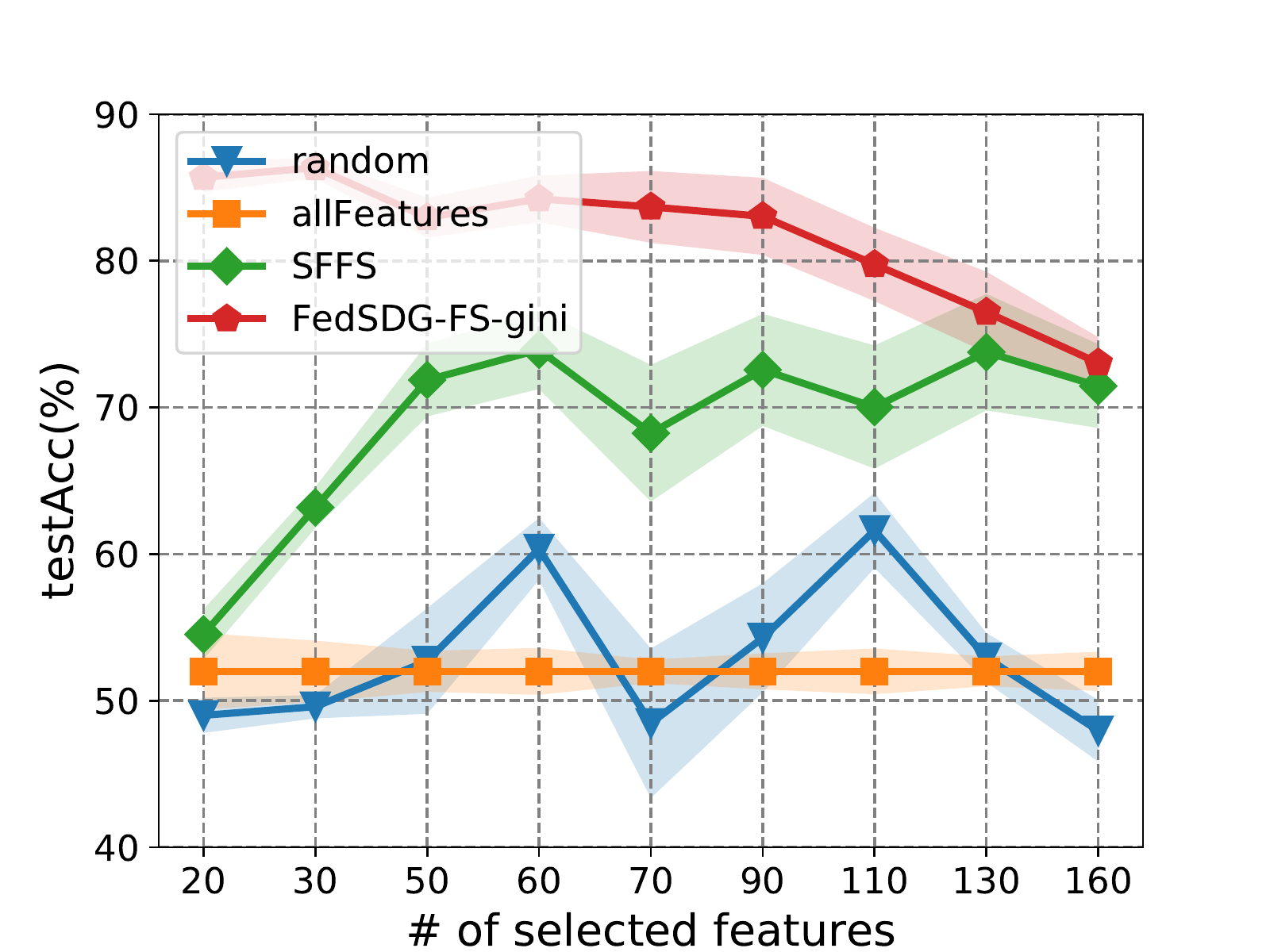}}	
		   \subfigure[\texttt{VFLLR-FRIEDMAN}]{\label{fig:friedman_gini_testAcc}
    		    \includegraphics[width = 0.48\linewidth, height=1.16in, trim=4 4 4 4]{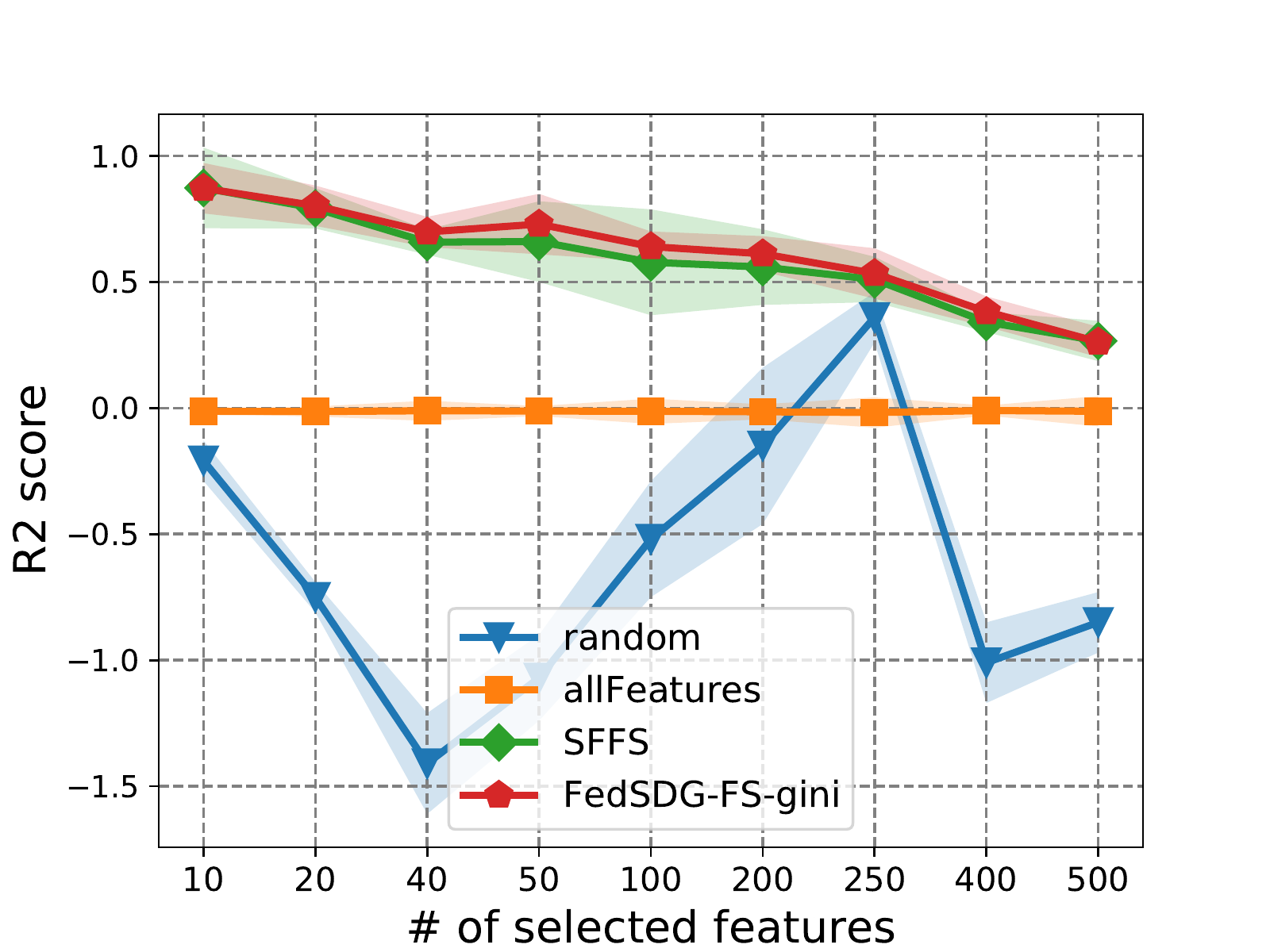}}
    		 \vspace{-0.1in}
    		\caption{Test accuracy and $R^2$ scores vs. number of selected features on synthetic datasets.}
    		\label{fig:testAcc_synthetic}
		\end{minipage}
		\quad
		\begin{minipage}[c]{0.48\linewidth}
		\centering
		\centerline{
    	\subfigure[\texttt{VFLNN-ARCENE}]{\label{fig:arcene_testAcc_fixedFeature}
    			    \includegraphics[width = 0.48\linewidth, height=1.16in, trim=4 4 4 4]{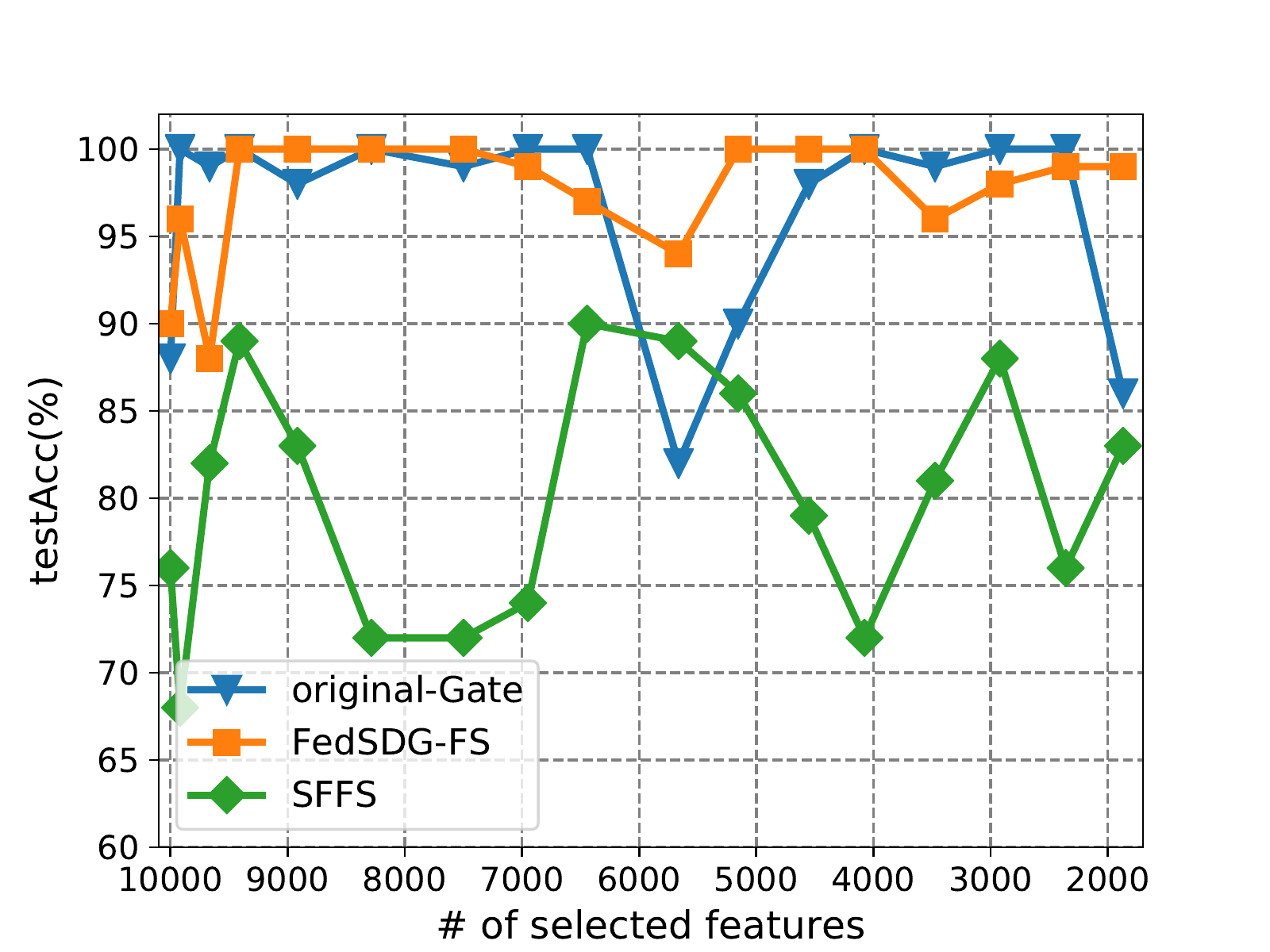}}	
	   \subfigure[\texttt{VFLNN-GISETTE}]{\label{fig:gisette_testAcc_fixedFeature}
		    \includegraphics[width = 0.48\linewidth, height=1.16in, trim=4 4 4 4]{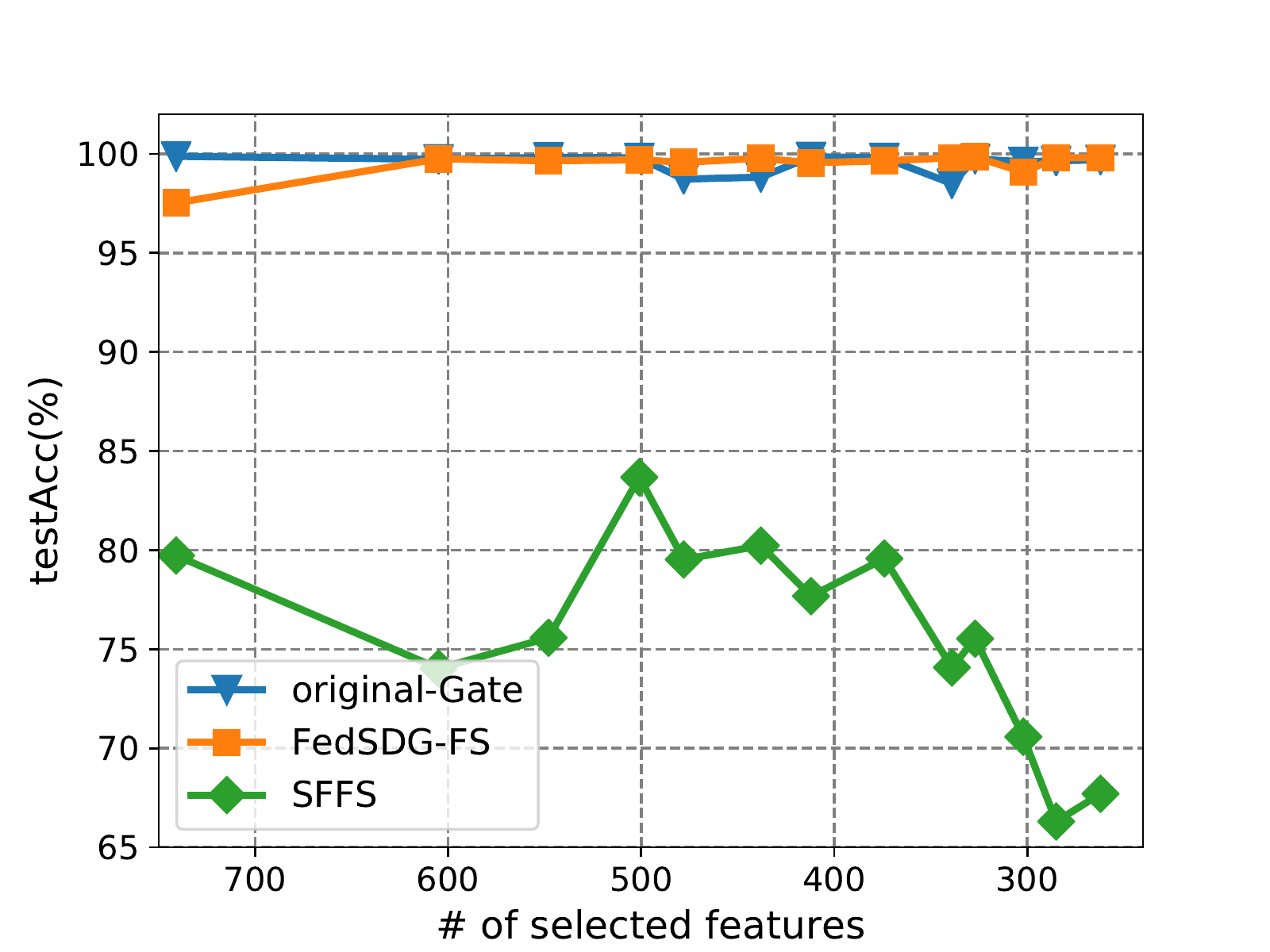}}
		    }
		   \vspace{-0.1in}
		\caption{Test accuracy vs. number of selected features by training models for 20 rounds.} 
	    \label{fig:testAcc_fixedFeature}
		\end{minipage}	 
	\end{center}
	\vspace{-0.12in}
\end{figure*}

\begin{figure*}[t!]
	\begin{center}
		\begin{minipage}[c]{0.97\linewidth}
			\subfigure[\texttt{VFLNN-ARCENE}]{\label{fig:arcene_testAcc_rounds}
			    \includegraphics[width = 0.236\linewidth, height=1.16in, trim=4 4 4 4]{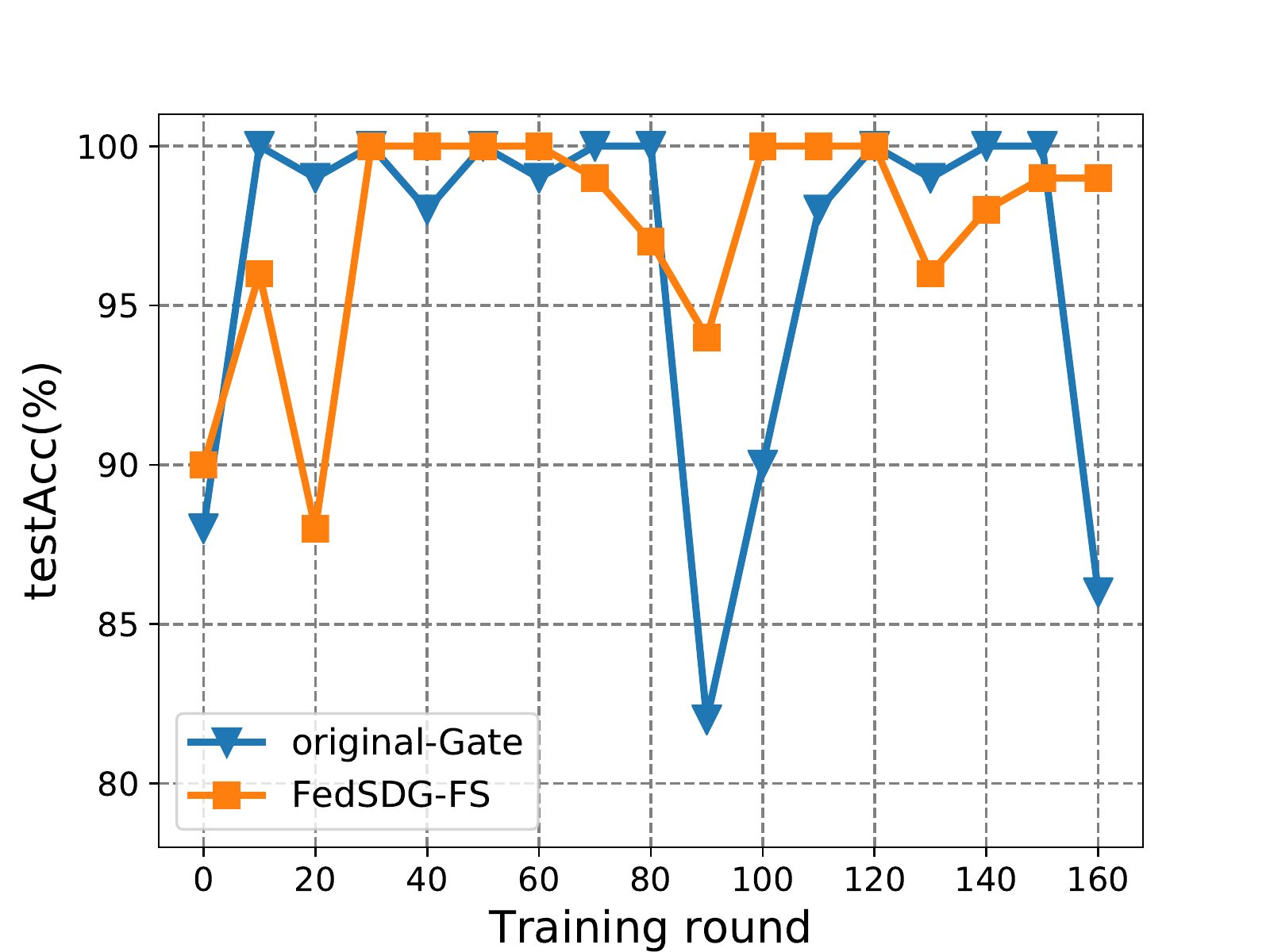}}	
		   \subfigure[\texttt{VFLNN-GISETTE}]{\label{fig:gisette_testAcc_rounds}
			    \includegraphics[width = 0.236\linewidth, height=1.16in, trim=4 4 4 4]{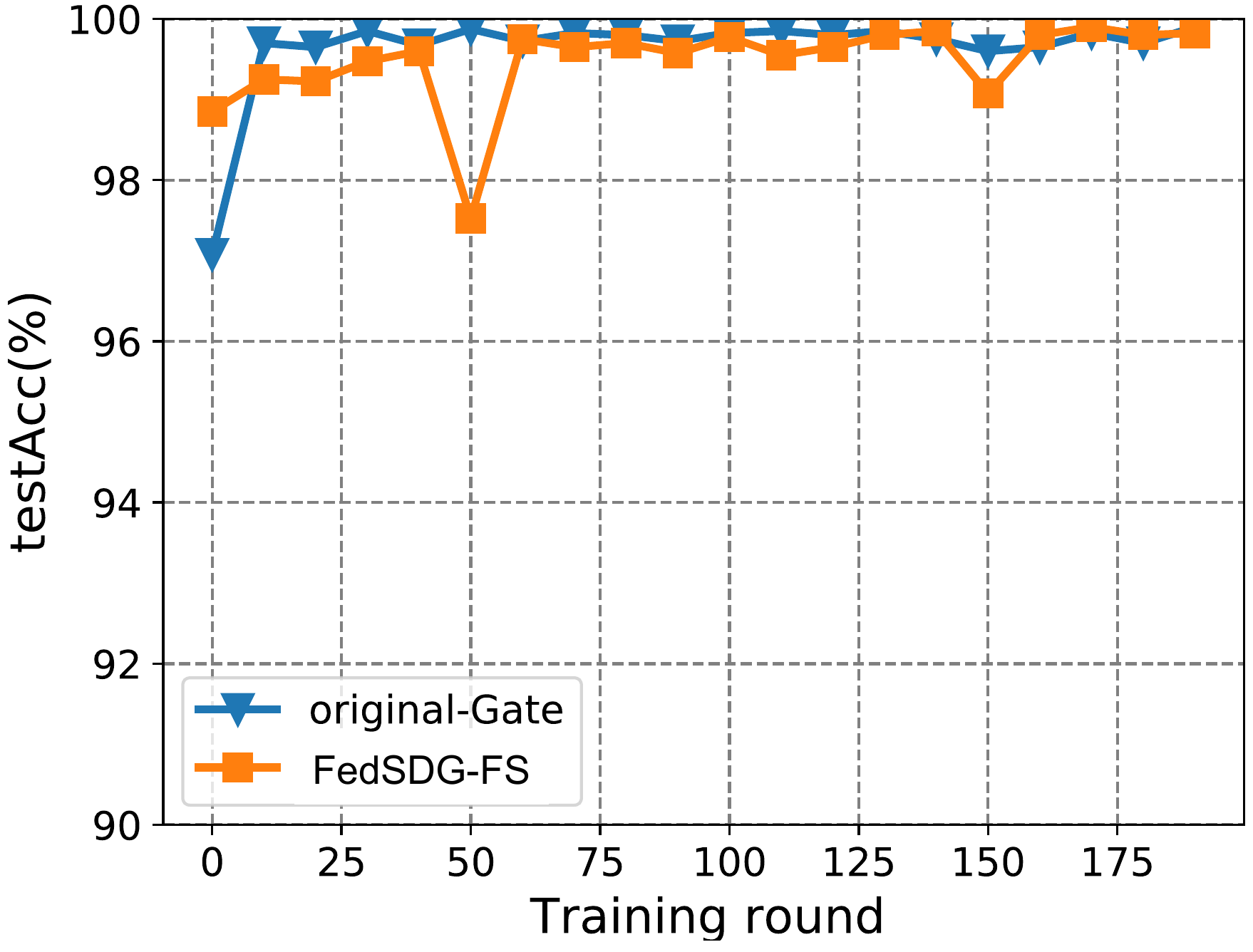}}
			\subfigure[\texttt{VFLNN-ARCENE}]{\label{fig:arcene_feature_rounds}
    			    \includegraphics[width = 0.236\linewidth, height=1.16in, trim=4 4 4 4]{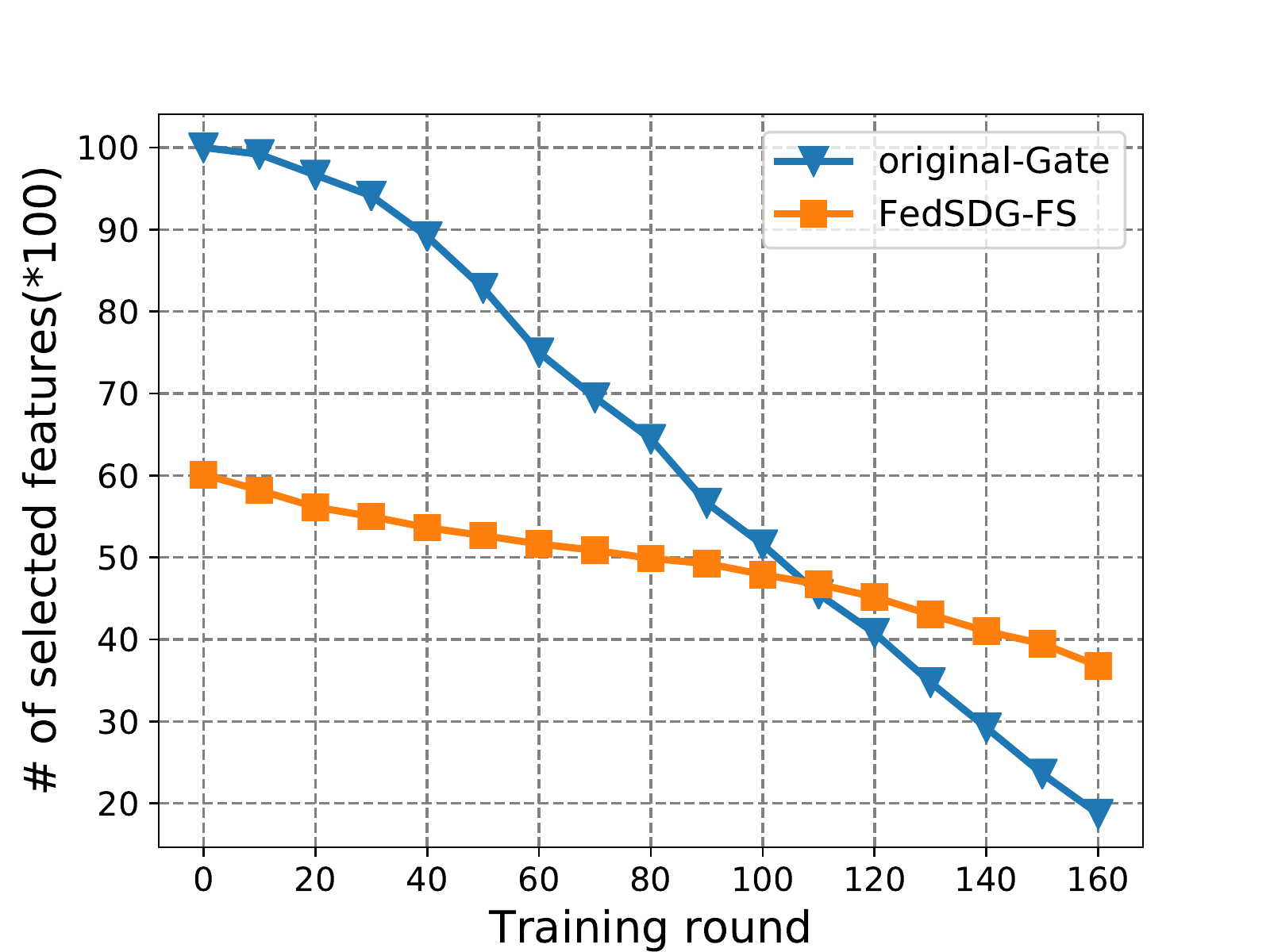}}	
	   \subfigure[\texttt{VFLNN-GISETTE}]{\label{fig:gisette_feature_rounds}
		    \includegraphics[width = 0.236\linewidth, height=1.16in, trim=4 4 4 4]{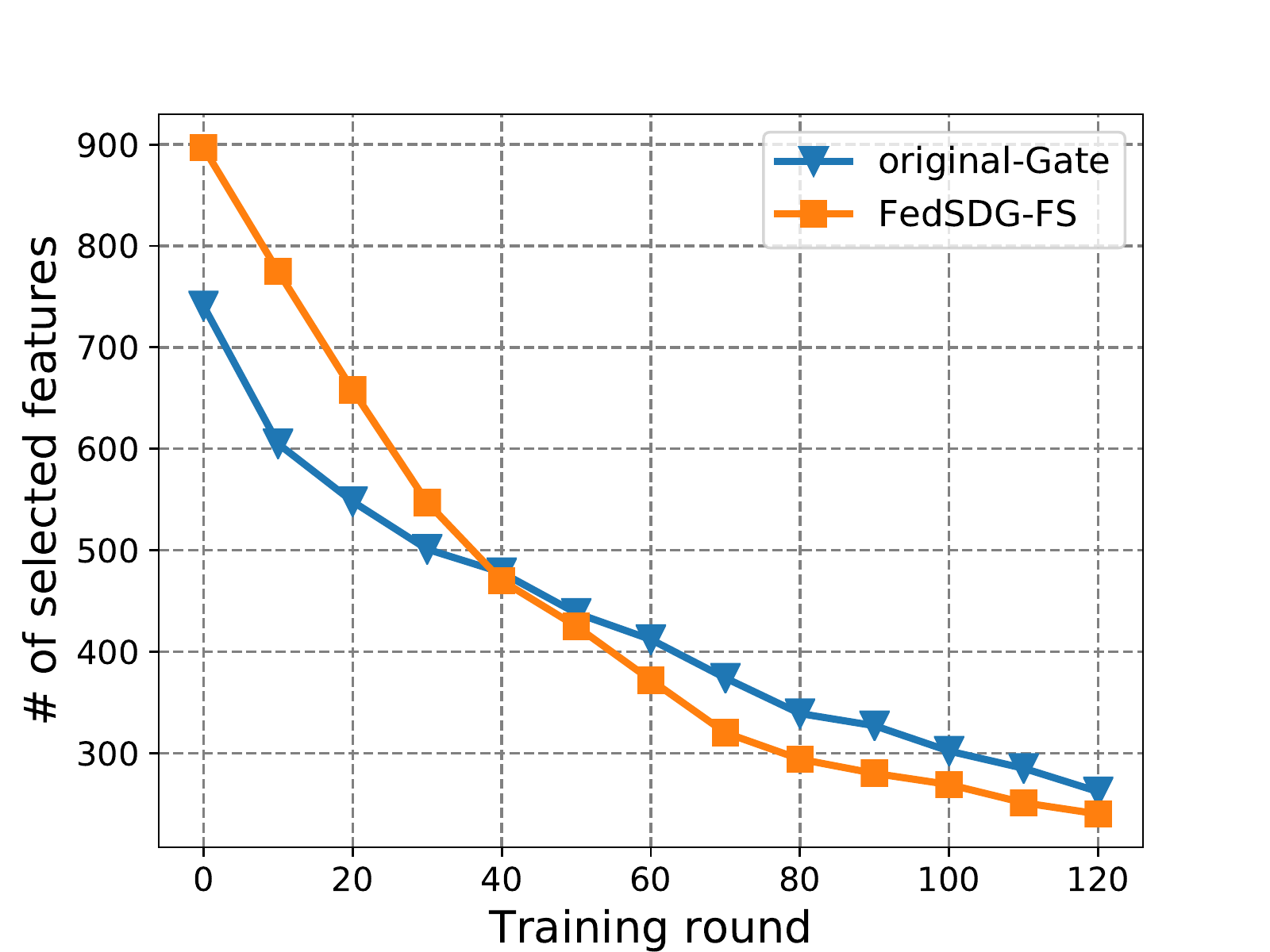}}
    		 \vspace{-0.1in}    		\caption{Test accuracy and number of selected features during model training of original gate selection method and FedSDG-FS.}
    		\label{fig:testAcc_Number_Feature}
		\end{minipage}
	\end{center}
	\vspace{-0.2in}
\end{figure*}
\vspace{-0.1in}
\subsection{Evaluating Gini Impurity for VFL}
Firstly, we evaluate the effectiveness of our Gini impurity metric (FedSDG-FS-gini) designed for feature importance initialization in FedSDG-FS by comparing the test accuracy of the global models to the other three filtering based feature selection strategies, SFFS \cite{pansecure}, random FS, and all features participating (allFeatures). 
% We employ two clients A and B, to jointly train a vertical neural networks based on \texttt{MADELONE}, and a vertical logistic regression model based on \texttt{FRIEDMAN}. 
We select different numbers of features (i.e., $k$ features with the smallest Gini scores), and assign them to the two clients. 
Since those datasets differ in both the number of features and the number of noisy features. Thus, we select features from each dataset in similar proportions and round the numbers of selected features.
We perform 5-fold cross validation and report the average $R^2$ scores, test accuracy and standard deviations in Fig. \ref{fig:testAcc_synthetic}.
Here, the $R^2$ score is defined as $R^2=1-\frac{\sum_{n\in [N]} (y_n-\hat{y}_n)^2}{\sum_{n\in [N]} (y_n-\bar{y})}$, where $\bar{y}=\frac{1}{N}\sum_{n\in [N]} y_n$, and $y_n$, $\hat{y}_n$ are the true target and predicted target of the $n$-th sample, respectively. 
The results show that FedSDG-FS-gini achieves higher test accuracy and $R^2$ scores than other strategies. 
Specifically, the average test accuracy and $R^2$ scores of FedSDG-FS-gini are 28.71\%/ 123.8\%, 29.69\%/ 70.3\%, 12.85\%/ 3.4\% higher than that of random, allFeatures and SFFS for \texttt{MADELON} and \texttt{FRIEDMAN}, respectively.  Meanwhile, the standard deviations are relative small, \emph{e.g.}, with 1.22\% and 4.3\% smaller than that of SFFS for \texttt{MADELON} and \texttt{FRIEDMAN}.  
% 30.2\% and 7.5\% higher than that of allFeatures for datasets \texttt{MADELON} and \texttt{ARCENE}. The average $R^2$ score of FedSDG-FS is 70.3\% higher than that of allFeatures for dataset \texttt{FRIEDMAN}. Meanwhile, we see that the standard deviations are relative small, with 1.22\% and 4.3\% smaller than that of SFFS.  
Besides, the reason why some of the test accuracy in Fig \ref{fig:testAcc_synthetic}. is less than 50\% is that there are noisy features irrelevant to the learning task and a large number of redundant features possessed by local clients, samples with very similar or the same features may have completely opposite labels. 
% Such low quality training data makes the models perform almost like random guesses or slightly worse when they overfit the noise.

% The $R^2$ score of the trained model \texttt{VFLLR-FRIEDMAN} and test accuracy of \texttt{VFLNN-MADELON} and \texttt{VFLNN-ARCENE} models on test datasets are shown in Figure \ref{fig:friedman_gini_testAcc}-\ref{fig:ARCENE_Gini1}. 
% For the accuracy validation, we further consider the precision/ recall of the method and test accuracy of different models trained using the features selected with this method.  
% \textbf{Learning Accuracy.} 

% \textbf{Efficiency.} Fig. gives the cost of 

% We measure the effectiveness of the noisy feature filtering method from two aspects. First, considering that we care about whether the identified noisy features are indeed noisy features, we use the indicator precision to characterize the effectiveness of the filtering method, where the precision is defined as the ratio of the features identified as noisy features that are actually noisy. We calculate for all individual features, and filter those with large Gini scores. 

\subsection{Evaluating Important Feature Selection} 
After feature importance initialization, FedSDG-FS proceeds to select important features using the stochastic dual-gates. 
We now evaluate our FedSDG-FS method compared to other baselines, all features participating (allFeatures),
SFFS, VFLFS \cite{louizos2017learning}, and the original gate based method which has neither gates of the embedding vectors nor importance initialization (original-Gate), using various datasets. For fair comparison, we implement VFLFS  \cite{louizos2017learning} without the part that makes use of the non-overlapping samples.
Further, we extend a filter feature selection method MS-GINI \cite{li2021privacy} based on Gini impurity in VFL settings to compare with FedSDG-FS. 
We perform 5-fold cross validation and report average accuracy. 

\begin{figure}[t!]
	\begin{minipage}[t]{0.464\linewidth}
		\centering
		\includegraphics[width=\linewidth, height=1.18in, trim=4 4 4 4]{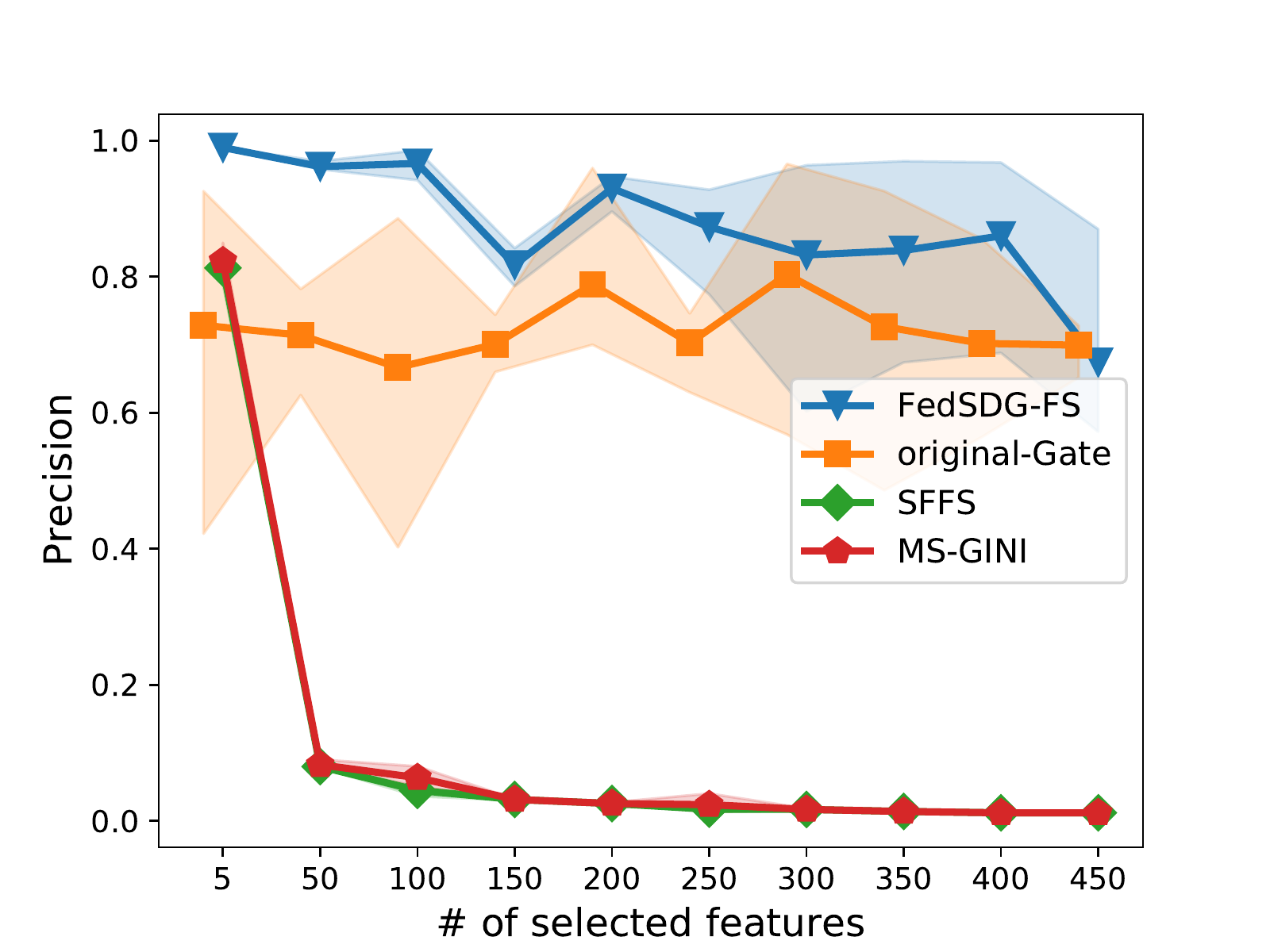}
		\vspace{-0.16in}
		\caption{Precision of different methods on \texttt{MADELON}.}
			\vspace{-0.12in}	
		\label{fig:precision_stg_sffs}
	\end{minipage}
	\quad
	\begin{minipage}[t]{0.464\linewidth}
 		\centering
		\includegraphics[width=\linewidth, height=1.18in, trim=4 4 4 4]{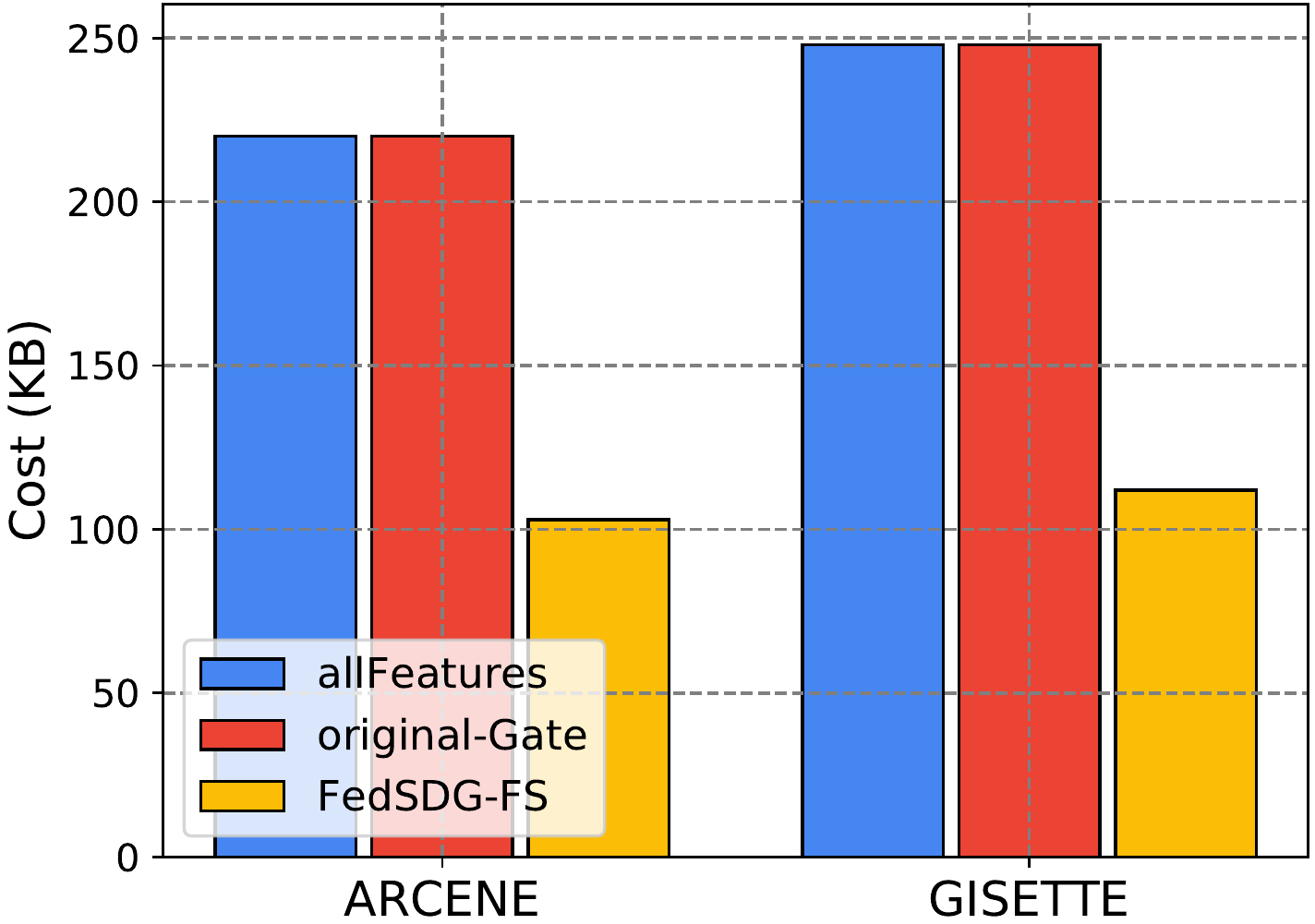}
		\vspace{-0.16in}
		\caption{Communication cost for 100 samples.}
			\vspace{-0.12in}	
		\label{fig:communication_cost}
	\end{minipage}
	%\end{center}
	\vspace{-0.1in}
\end{figure}

\textbf{1) Precision.}
% \subsubsection{\textbf{Precision}} 
We use precision to measure the accuracy of FedSDG-FS, which calculates the proportion of correctly selected informative features over all selected features. For FedSDG-FS and the original gate method, we train \texttt{VFLNN-MADELONE} until the model converges, and determine the important features. For SFFS and MS-GINI, we calculate the F-statistics and Gini impurity of each individual feature, respectively, and select different numbers of informative features. 
The results are shown in Fig \ref{fig:precision_stg_sffs}. It can be observed that FedSDG-FS and the original gate method achieve much higher precision than allFeatures, SFFS, MS-GINI, and \textbf{FedSDG-FS achieves the highest precision}. This illustrates that reducing the sizes of embedding vectors does not degrade the model accuracy.
For example, the average precision scores of different number settings of FedSDG-FS are 13\% and 76\% higher than the original gate method and SFFS, respectively.
As the number of selected features increases, the precision of SFFS and MS-GINI decreases dramatically, while the presicion decreases slightly for the FedSDG-FS and original gate methods, which demonstrates the effectiveness of FedSDG-FS without knowing the number of features to be selected.

% \subsubsection{\textbf{Learning Accuracy}}
\textbf{2) Learning Accuracy.}
We compare FedSDG-FS with others by training different VFL models and evaluating the test accuracy of the global models, and the ratios of selected features. The results are shown in Table \ref{tab:testAcc_dataset}. It can be observed that \textbf{FedSDG-FS achieves the highest test accuracy using the fewest features in almost all datasets}. Taking \texttt{MADELON} as an example, the average test accuracy of FedSDG-FS is 0.3\%, 33.6\%, 27.0\%, 47.2\%, 50.2\% higher than the four methods; while the ratio of selected features is 0.02, 0.97, 0.47, 0.47, 0.47 less than them. 
% For the only case where FedSDG-FS performs second best in terms of accuracy on dataset \texttt{PCMAC}, our accuracy 98.7\% is much close to the best accuracy 99.1\%, while FedSDG-FS use about 20\% fewer features. 
In some cases where there are small number of noisy features, and having little negative impact on the model, using all features results the higher accuracy. Nevertheless, FedSDG-FS can still achieve comparable test accuracy with fewer features. To further validate the proposed methods, we conducted experiments with five clients and ten clients. Two example results are presented in Table \ref{tab:testAcc_moreClients}, which show that FedSDG-FS achieves the highest test accuracy using the fewest features in most cases. For the only case where FedSDG-FS performs second best in terms of accuracy, our accuracy 99.5\% is very close to the best accuracy 99.8\%, while FedSDG-FS use about 20\% fewer features. 
	
\begin{table}[ht!]
	\centering
	\vspace{-0.1in}
	\caption{Test accuracy of models trained with features selected.} 
	\vspace{-0.1in}
	\label{tab:testAcc_dataset}
	\resizebox{1\linewidth}{!}
	{
		\begin{tabular}{l| c| c|c|c|c| l}
			\toprule 
			\textbf{Datasets} & \multicolumn{5}{c}{\textbf{Test Accuracy (\%) / Ratio of Selected Features}} \\
			\hline	&allFeatures&SFFS &MS-GINI&VFLFS&original-Gate& \textbf{FedSDG-FS}\\
			\hline  
			\texttt{MADELON} & 52.0/ 1.0 & 65.6/ 0.5&72.2/ 0.5 &51.0/ 0.5 &98.9/ 0.05 &\textbf{99.2}/ 0.03
			 \\
			 \hline  \texttt{ARCENE} & 80.1/ 1.0& 95.0/ 0.5& 87.5/ 0.5 &70.1/ 0.5 &97.4/ 1.0 &\textbf{99.8}/ 0.58
			 \\
			 \hline \texttt{BASEHOCK} &99.7/ 1.0 &99.1/ 0.5 &98.5/ 0.5 &94.4/ 0.5  &99.5/ 0.48 &\textbf{99.9}/ 0.3 \\
			 \hline \texttt{RELATHE} &95.5/ 1.0 &87.2/ 0.5 &92.1/ 0.5 &86.5/ 0.5&99.7/ 0.71 & \textbf{99.8}/ 0.41\\
			 \hline \texttt{PCMAC} &97.6/ 1.0&79.34/ 0.5 & 90.2/ 0.5 &86.11/ 0.5 &\textbf{99.1}/ 0.66 &98.7/ 0.45 \\
			 \hline \texttt{GISETTE} &99.1/ 1.0&99.0/ 0.5 &98.0/ 0.5 &50.2/ 0.5 &99.3/ 0.81 & \textbf{99.5}/ 0.53 \\
			 \hline \texttt{COIL20} &96.4/ 1.0&65.6/ 0.5&94.8/ 0.5&72.2/ 0.5 & 91.2/ 1.0 & \textbf{97.5}/ 0.71 \\
			 \hline \texttt{ISOLET} &\textbf{98.0}/ 1.0& 92.7/ 0.5&91.4/ 0.5&71.4/ 0.5&93.2/ 1.0 &96.7/ 0.75\\
			\toprule
		\end{tabular}
		\vspace{-0.2in}
	}
\end{table}
\begin{table}[ht!]
	\centering	
	\vspace{-0.2in}
	\caption{Test accuracy of models trained with features selected.} 
	\vspace{-0.1in}
	\label{tab:testAcc_moreClients}
	\resizebox{1\linewidth}{!}
	{
		\begin{tabular}{l |c| c|c|c|c| l}
			\toprule 
			\textbf{Datasets} & \multicolumn{6}{c}{\textbf{Test Accuracy (\%) / Ratio of Selected Features}} \\
		\hline	 \multicolumn{7}{c}{\textbf{5 Clients}} \\
			\hline	&allFeatures&SFFS &MS-GINI& VFLFS&original-Gate& \textbf{FedSDG-FS}\\
			\hline  
			\texttt{ARCENE} & 85.8/ 1.0 & 92.2/ 0.5&92.0/ 0.5 &71.0/ 0.5 &98.2/ 1.0 &\textbf{99.7}/ 0.59
			 \\
			 \hline  \texttt{RELATHE} & 97.6/ 1.0& 84.8/ 0.5& 92.7/ 0.5& 86.1/ 0.5 &\textbf{99.8}/ 0.69 & 99.5/ 0.45
			 \\
			 \hline	 \multicolumn{7}{c}{\textbf{10 Clients}} \\
			\toprule	&allFeatures&SFFS &MS-GINI&VFLFS&original-Gate& \textbf{FedSDG-FS}\\
			\hline  
			\texttt{ARCENE} & 91.0/ 1.0 & 94.0/ 0.5&92.7/ 0.5 &82.0/ 0.5 &98.6/ 1.0 &\textbf{99.2}/ 0.56
			 \\
			 \hline  \texttt{RELATHE} & 97.5/ 1.0& 85.6/ 0.5& 93.6/ 0.5 & 85.7/ 0.5& 99.5/ 0.68 &\textbf{99.8}/ 0.44
			 \\
			\hline
		\end{tabular}
		\vspace{-0.1in}
	}
\end{table}
% \subsubsection{\textbf{Stability}} 
\textbf{3) Stability.}
We evaluate the stability of FedSDG-FS from two aspects, 1) test accuracy of the global model with different numbers of selected features, and 2) test accuracy at different training rounds. 
We illustrate the test accuracy of models \texttt{VFLNN-ARCENE} and \texttt{VFLNN-GISETTE} by training them for 20 rounds with different numbers of features in Fig. \ref{fig:testAcc_fixedFeature}. 
The results show that compared to SFFS, the original based method and FedSDG-FS both achieve much higher test accuracy. The performance of FedSDG-FS has little variation in all cases. Then, we calculate the test accuracy of the two models in different training rounds, and plot them in Fig. \ref{fig:arcene_testAcc_rounds} and Fig. \ref{fig:gisette_testAcc_rounds}.  The results show that FedSDG-FS and the original gate method achieve comparably high test accuracies at different training rounds, while FedSDG-FS is more stable (i.e., the test accuracy of global model drops 2.8\% in the 80-th round for FedSDG-FS and 17.9\% for the original gate method). The analysis results of test accuracy on other models with different numbers of selected features and at different training rounds are similar to that of \texttt{VFLNN-ARCENE}, \texttt{VFLNN-GISETTE}. 
% See the appendix for detailed results on other datasets. 

% \subsubsection{\textbf{Efficiency}} 
\textbf{4) Efficiency.}
Finally, we evaluate the efficiency of FedSDG-FS from two aspects: 1) the speed of the feature importance initialization, and 2) communication saving during model prediction. 
Firstly, we calculate the number of selected features in different rounds of training \texttt{VFLNN-ARCENE} and \texttt{VFLNN-GISETTE} (Fig. \ref{fig:arcene_feature_rounds} and Fig. \ref{fig:gisette_feature_rounds}). The results show that with importance initialization, the models can quickly filter out noisy features and select important ones, thus speeding up model training. 
Secondly, we compare the prediction communication overhead of those models of FedSDG-FS, allFeatures and the original gate method. Fig. \ref{fig:communication_cost} shows the average communication cost of each method to select features. The communication cost of FedSDG-FS is more than 50\% lower than that of the other methods (\emph{e.g.}, 53.2\%, 54.7\% lower for datasets \texttt{ARCENE} and \texttt{GISETTE}). The efficiency analysis clearly demonstrated the advantages of the feature importance initialization module of FedSDG-FS.

\section{Conclusions}
\label{sec:conclusion}
In this work, we proposed an efficient and secure vertical federated learning feature selection framework to select important features in VFL settings.
We first designed a Gaussian stochastic dual-gates for clients' inputs to efficiently approximate the probability of a feature being selected. 
Then, we incorporated PHE and randomized noise mechanism into stochastic dual-gates to achieve secure feature selection. 
To reduce overhead, we proposed a feature importance initialization method based on Gini impurity and PHE, which can be accomplished through only two parameter transmissions, and two encryption/decryption operation on the server. 
Experiment results show that FedSDG-FS significantly outperforms existing approaches in terms of achieving more accurate selection of high-quality features and building global models with better performance.
% The noisy feature identification step can be accomplished through a single parameter transmission between the VFL server and the clients, and only one encryption/decryption operation on the server based on Gini impurity and PHE. The secure important feature selection method leverages Gaussian-based stochastic gates for VFL, and is designed based on PHE and the randomized noise mechanism to simultaneously determine important features and construct the global model for enhanced performance. 
FedSDG-FS achieves the privacy protection goal, \emph{e.g.}, during the entire feature selection and model training process, neither data nor labels will be acquired or inferred by any party other than their original owners. 
% In subsequent research, we aim to enhance FedSDG-FS to distinguish more diverse types of noisy features (\emph{e.g.}, shortcut features \cite{geirhos2020shortcut}) in order to withstand inference attacks from malicious clients.

\section*{Acknowledgments}
Han Yu is the corresponding author. This research is supported by Nanyang Technological University (NTU), under SUG Grant (020724-00001); the National Research Foundation, Prime Ministers Office, National Cybersecurity R\&D Program (No. NRF2018NCR-NCR005-0001), NRF Investigatorship NRF-NRFI06-2020-0001; the National Research Foundation, Singapore and DSO National Laboratories under the AI Singapore Programme (AISG Award No: AISG2-RP-2020-019); Alibaba Group through Alibaba Innovative Research (AIR) Program and Alibaba-NTU Singapore Joint Research Institute (JRI) (Alibaba-NTU-AIR2019B1), NTU, Singapore; the RIE 2020 Advanced Manufacturing and Engineering Programmatic Fund (No. A20G8b0102), Singapore; NTU Nanyang Assistant Professorship, Future Communications Research \& Development Programme (FCP-NTU-RG-2021-014), the National Key R\&D Program of China 2021YFB2900103, China National Natural Science Foundation with No. 61932016, and ``the Fundamental Research Funds for the Central Universities" WK2150110024. 

% \newpage

\bibliographystyle{IEEEtran}
\bibliography{reference}

\begin{thebibliography}{10}
\providecommand{\url}[1]{#1}
\csname url@samestyle\endcsname
\providecommand{\newblock}{\relax}
\providecommand{\bibinfo}[2]{#2}
\providecommand{\BIBentrySTDinterwordspacing}{\spaceskip=0pt\relax}
\providecommand{\BIBentryALTinterwordstretchfactor}{4}
\providecommand{\BIBentryALTinterwordspacing}{\spaceskip=\fontdimen2\font plus
\BIBentryALTinterwordstretchfactor\fontdimen3\font minus
  \fontdimen4\font\relax}
\providecommand{\BIBforeignlanguage}[2]{{%
\expandafter\ifx\csname l@#1\endcsname\relax
\typeout{** WARNING: IEEEtran.bst: No hyphenation pattern has been}%
\typeout{** loaded for the language `#1'. Using the pattern for}%
\typeout{** the default language instead.}%
\else
\language=\csname l@#1\endcsname
\fi
#2}}
\providecommand{\BIBdecl}{\relax}
\BIBdecl

\bibitem{mcmahan2017communication}
B.~McMahan, E.~Moore, D.~Ramage, S.~Hampson, and B.~A. y~Arcas,
  ``Communication-efficient learning of deep networks from decentralized
  data,'' in \emph{Artificial intelligence and statistics}.\hskip 1em plus
  0.5em minus 0.4em\relax PMLR, 2017, pp. 1273--1282.

\bibitem{hu2019fdml}
Y.~Hu, D.~Niu, J.~Yang, and S.~Zhou, ``Fdml: A collaborative machine learning
  framework for distributed features,'' in \emph{Proceedings of the 25th ACM
  SIGKDD International Conference on Knowledge Discovery \& Data Mining}, 2019,
  pp. 2232--2240.

\bibitem{yang2019federated}
Q.~Yang, Y.~Liu, Y.~Cheng, Y.~Kang, T.~Chen, and H.~Yu, ``Federated learning,''
  \emph{Synthesis Lectures on Artificial Intelligence and Machine Learning},
  vol.~13, no.~3, pp. 1--207, 2019.

\bibitem{wang2022efficient}
J.~Wang, L.~Zhang, A.~Li, X.~You, and H.~Cheng, ``Efficient participant
  contribution evaluation for horizontal and vertical federated learning,'' in
  \emph{2022 IEEE 38th International Conference on Data Engineering
  (ICDE)}.\hskip 1em plus 0.5em minus 0.4em\relax IEEE, 2022, pp. 911--923.

\bibitem{li2021privacy1}
A.~Li, L.~Zhang, J.~Wang, F.~Han, and X.-Y. Li, ``Privacy-preserving efficient
  federated-learning model debugging,'' \emph{IEEE Transactions on Parallel and
  Distributed Systems}, vol.~33, no.~10, pp. 2291--2303, 2021.

\bibitem{li2021efficient}
A.~Li, L.~Zhang, J.~Wang, J.~Tan, F.~Han, Y.~Qin, N.~M. Freris, and X.-Y. Li,
  ``Efficient federated-learning model debugging,'' in \emph{2021 IEEE 37th
  International Conference on Data Engineering (ICDE)}.\hskip 1em plus 0.5em
  minus 0.4em\relax IEEE, 2021, pp. 372--383.

\bibitem{zhuang2021joint}
W.~Zhuang, Y.~Wen, and S.~Zhang, ``Joint optimization in edge-cloud continuum
  for federated unsupervised person re-identification,'' in \emph{Proceedings
  of the 29th ACM International Conference on Multimedia}, 2021, pp. 433--441.

\bibitem{li2021sample}
A.~Li, L.~Zhang, J.~Tan, Y.~Qin, J.~Wang, and X.-Y. Li, ``Sample-level data
  selection for federated learning,'' in \emph{IEEE INFOCOM 2021-IEEE
  Conference on Computer Communications}.\hskip 1em plus 0.5em minus
  0.4em\relax IEEE, 2021, pp. 1--10.

\bibitem{liu2019communication}
Y.~Liu, Y.~Kang, L.~Li, X.~Zhang, Y.~Cheng, T.~Chen, M.~Hong, and Q.~Yang, ``A
  communication efficient vertical federated learning framework,''
  \emph{Unknown Journal}, 2019.

\bibitem{chen2020vafl}
T.~Chen, X.~Jin, Y.~Sun, and W.~Yin, ``Vafl: a method of vertical asynchronous
  federated learning,'' \emph{arXiv preprint arXiv:2007.06081}, 2020.

\bibitem{tan2022residue}
J.~Tan, L.~Zhang, Y.~Liu, A.~Li, and Y.~Wu, ``Residue-based label protection
  mechanisms in vertical logistic regression,'' \emph{arXiv preprint
  arXiv:2205.04166}, 2022.

\bibitem{PowerFL}
PowerFL, ``Angel powerfl,'' \url{https://data.qq.com/powerfl/}.

\bibitem{FATE}
FATE, ``Fate-federated-ai,'' \url{https://github.com/FederatedAI/DOC-CHN}.

\bibitem{li2021privacy}
X.~Li, R.~Dowsley, and M.~De~Cock, ``Privacy-preserving feature selection with
  secure multiparty computation,'' \emph{ICML 2021}, 2021.

\bibitem{yamada2020feature}
Y.~Yamada, O.~Lindenbaum, S.~Negahban, and Y.~Kluger, ``Feature selection using
  stochastic gates,'' in \emph{International Conference on Machine
  Learning}.\hskip 1em plus 0.5em minus 0.4em\relax PMLR, 2020, pp.
  10\,648--10\,659.

\bibitem{chen2017kernel}
J.~Chen, M.~Stern, M.~J. Wainwright, and M.~I. Jordan, ``Kernel feature
  selection via conditional covariance minimization,'' \emph{NeurIPS 2017},
  2017.

\bibitem{pansecure}
F.~Pan, D.~Meng, Y.~Zhang, H.~Li, and X.~Li, ``Secure federated feature
  selection for cross-feature federated learning,'' 2020.

\bibitem{song2012feature}
L.~Song, A.~Smola, A.~Gretton, J.~Bedo, and K.~Borgwardt, ``Feature selection
  via dependence maximization.'' \emph{Journal of Machine Learning Research},
  vol.~13, no.~5, 2012.

\bibitem{estevez2009normalized}
P.~A. Est{\'e}vez, M.~Tesmer, C.~A. Perez, and J.~M. Zurada, ``Normalized
  mutual information feature selection,'' \emph{IEEE Transactions on neural
  networks}, vol.~20, no.~2, pp. 189--201, 2009.

\bibitem{roy2015feature}
D.~Roy, K.~S.~R. Murty, and C.~K. Mohan, ``Feature selection using deep neural
  networks,'' in \emph{2015 International Joint Conference on Neural Networks
  (IJCNN)}.\hskip 1em plus 0.5em minus 0.4em\relax IEEE, 2015, pp. 1--6.

\bibitem{kabir2010new}
M.~M. Kabir, M.~M. Islam, and K.~Murase, ``A new wrapper feature selection
  approach using neural network,'' \emph{Neurocomputing}, vol.~73, no. 16-18,
  pp. 3273--3283, 2010.

\bibitem{li2016deep}
Y.~Li, C.-Y. Chen, and W.~W. Wasserman, ``Deep feature selection: theory and
  application to identify enhancers and promoters,'' \emph{Journal of
  Computational Biology}, vol.~23, no.~5, pp. 322--336, 2016.

\bibitem{hans2009bayesian}
C.~Hans, ``Bayesian lasso regression,'' \emph{Biometrika}, vol.~96, no.~4, pp.
  835--845, 2009.

\bibitem{louizos2017learning}
C.~Louizos, M.~Welling, and D.~P. Kingma, ``Learning sparse neural networks
  through $l_0$ regularization,'' \emph{arXiv preprint arXiv:1712.01312}, 2017.

\bibitem{zhang2020additively}
Y.~Zhang and H.~Zhu, ``Additively homomorphical encryption based deep neural
  network for asymmetrically collaborative machine learning,'' \emph{arXiv
  preprint arXiv:2007.06849}, 2020.

\bibitem{cheng2021secureboost}
K.~Cheng, T.~Fan, Y.~Jin, Y.~Liu, T.~Chen, D.~Papadopoulos, and Q.~Yang,
  ``Secureboost: A lossless federated learning framework,'' \emph{IEEE
  Intelligent Systems}, vol.~36, no.~6, pp. 87--98, 2021.

\bibitem{song2007supervised}
L.~Song, A.~Smola, A.~Gretton, K.~M. Borgwardt, and J.~Bedo, ``Supervised
  feature selection via dependence estimation,'' in \emph{Proceedings of the
  24th international conference on Machine learning}, 2007, pp. 823--830.

\bibitem{allen2013automatic}
G.~I. Allen, ``Automatic feature selection via weighted kernels and
  regularization,'' \emph{Journal of Computational and Graphical Statistics},
  vol.~22, no.~2, pp. 284--299, 2013.

\bibitem{feng2022vertical}
S.~Feng, ``Vertical federated learning-based feature selection with
  non-overlapping sample utilization,'' \emph{Expert Systems with
  Applications}, p. 118097, 2022.

\bibitem{ARCENE}
\BIBentryALTinterwordspacing
J.~Thomas, ``Mass spectrometric data.'' [Online]. Available:
  \url{https://www.openml.org/d/41157}
\BIBentrySTDinterwordspacing

\bibitem{guyon2004result}
I.~Guyon, S.~Gunn, A.~Ben-Hur, and G.~Dror, ``Result analysis of the nips 2003
  feature selection challenge,'' \emph{Advances in neural information
  processing systems}, vol.~17, 2004.

\bibitem{miller2017reducing}
A.~Miller, N.~Foti, A.~D'Amour, and R.~P. Adams, ``Reducing reparameterization
  gradient variance,'' \emph{Advances in Neural Information Processing
  Systems}, vol.~30, 2017.

\bibitem{erkin2009privacy}
Z.~Erkin, M.~Franz, J.~Guajardo, S.~Katzenbeisser, I.~Lagendijk, and T.~Toft,
  ``Privacy-preserving face recognition,'' in \emph{International symposium on
  privacy enhancing technologies symposium}.\hskip 1em plus 0.5em minus
  0.4em\relax Springer, 2009, pp. 235--253.

\bibitem{friedman1991multivariate}
J.~H. Friedman, ``Multivariate adaptive regression splines,'' \emph{The annals
  of statistics}, vol.~19, no.~1, pp. 1--67, 1991.

\bibitem{real-world-datasets}
\BIBentryALTinterwordspacing
A.~state university, ``Feature selection datasets,'' Public online, 2010.
  [Online]. Available:
  \url{https://jundongl.github.io/scikit-feature/OLD/datasets_old.html}
\BIBentrySTDinterwordspacing

\bibitem{gisette}
\BIBentryALTinterwordspacing
U.~machine~learning repository, ``Handwritten digit recognition problem.''
  [Online]. Available: \url{https://archive.ics.uci.edu/ml/datasets/Gisette}
\BIBentrySTDinterwordspacing

\bibitem{coil20}
\BIBentryALTinterwordspacing
C.~University, ``Image classification task.'' [Online]. Available:
  \url{https://www.cs.columbia.edu/CAVE/software/softlib/coil-20.php}
\BIBentrySTDinterwordspacing

\bibitem{isolet}
\BIBentryALTinterwordspacing
U.~machine~learning repository, ``Letter-name classification task.'' [Online].
  Available: \url{https://archive.ics.uci.edu/ml/datasets/isolet}
\BIBentrySTDinterwordspacing

\bibitem{hardy2017private}
S.~Hardy, W.~Henecka, H.~Ivey-Law, R.~Nock, G.~Patrini, G.~Smith, and
  B.~Thorne, ``Private federated learning on vertically partitioned data via
  entity resolution and additively homomorphic encryption,'' \emph{arXiv
  preprint arXiv:1711.10677}, 2017.

\bibitem{flower-FL}
\BIBentryALTinterwordspacing
Flower, ``Flower: A friendly federated learning framework,'' Public online,
  2022. [Online]. Available: \url{https://flower.dev/}
\BIBentrySTDinterwordspacing

\bibitem{Pytorch}
\BIBentryALTinterwordspacing
Pytorch, ``Pytorch,'' Public online, 2022. [Online]. Available:
  \url{https://pytorch.org/}
\BIBentrySTDinterwordspacing

\end{thebibliography}

\end{document}